\documentclass[lettersize,journal]{IEEEtran}

\newif\ifblind
\blindfalse

\usepackage{amsmath,amsfonts,amsthm,amssymb}
\usepackage{array}
\usepackage[caption=false,font=normalsize,labelfont=sf,textfont=sf]{subfig}
\usepackage{textcomp}
\usepackage{stfloats}
\usepackage{url}
\usepackage{verbatim}
\usepackage{graphicx}
\usepackage{cite}
\usepackage{bm}
\usepackage{float}
\usepackage{hyperref}
\usepackage{xcolor}
\usepackage{multirow}
\usepackage{booktabs}
\usepackage[export]{adjustbox}
\usepackage{tikz}
\usetikzlibrary{arrows.meta,positioning,shapes.geometric}
\usepackage{todonotes}
\newcommand{\jiayurev}[1]{{\color{black}#1 }}

\def\BibTeX{{\rm B\kern-.05em{\sc i\kern-.025em b}\kern-.08em
    T\kern-.1667em\lower.7ex\hbox{E}\kern-.125emX}}

\DeclareMathOperator*{\argmax}{arg\,max}

\begin{document}

\hypersetup{pdfauthor={Jiayu Huang, Boyang Xu, Yongming Liu, Hao Yan},pdftitle={Uncertainty-Aware Joint Generative Modeling of Fault Event Sequences and Sensor Time Series in Solar PV Plants}}

\title{Uncertainty-Aware Joint Generative Modeling of Fault Event Sequences and Sensor Time Series in Solar PV Plants}

\author{Jiayu Huang$^{1}$, Boyang Xu$^{1}$, Yongming Liu$^{1}$, Hao Yan$^{1,2}$%
\thanks{$^{1}$Arizona State University, Tempe, AZ, USA}%
\thanks{$^{2}$Corresponding Author, \tt\small haoyan@asu.edu}%
}
\markboth{IEEE Transactions on Automation Science and Engineering}%
{Huang \MakeLowercase{\textit{et al.}}: Multi-modal Generative Modeling of Event Sequences and Time Series}

\maketitle
\begin{center}\footnotesize\textit{Author preprint. This work has been submitted to the IEEE for possible publication. Copyright may be transferred without notice, after which this version may no longer be accessible. Supplementary material is included as an appendix.}\end{center}

\begin{abstract}
This paper presents a multimodal generative model designed to simulate a solar energy plant system for predictive maintenance and fault prediction. We propose a novel approach to enhance system generation through conditional generation by leveraging both event sequences and time-series data and improve next event prediction. The framework employs a Transformer Hawkes Process (THP) to encode event data and an iTransformer to encode time-series data, integrated via a co-attention mechanism that effectively captures cross-modal dependencies between discrete events and continuous signals. From these integrated representations, we employ two specialized decoders for simulation: a Conditional Latent Diffusion Model (CLDM) for modeling future sensor trajectories and the Score Matching-based UnceRtainty quantiFication for Transformer Hawkes Process (SMURF-THP) for high-fidelity event prediction. \jiayurev{This dual-decoder strategy supports uncertainty-aware, history-conditioned simulation by rolling the model forward for a specified number of generation iterations, producing finite stochastic scenarios rather than deterministic extrapolations over the full observational record.} Evaluated on a comprehensive dataset from the Red Rock Solar Site in Arizona, our approach improves aggregate event-prediction F1 relative to baselines and supports applications in predictive maintenance, fault detection, and energy optimization for solar power systems.
\end{abstract}

\def\abstractname{Note to Practitioners}
\begin{abstract}
Photovoltaic (PV) plant operators face challenges in anticipating rare, multiclass faults from event logs alone. This paper introduces a multimodal generative model that combines sensor time-series data with fault event sequences to predict system behavior and future fault occurrences. The model employs generative decoders that provide probabilistic forecasts, enabling risk-sensitive maintenance planning. \jiayurev{Practitioners can use the model to generate finite stochastic operating scenarios with a specified number of rollout iterations, forecast near-term fault timing and type, and inform proactive maintenance scheduling without interpreting the output as deterministic two-year extrapolation.} The framework is demonstrated on data from the Red Rock Solar Site and can serve as a digital twin for evaluating maintenance policies safely via offline RL and off-policy evaluation before field trials. \jiayurev{Key practical benefits include improved prediction for several low-frequency fault classes with sufficient historical support, uncertainty estimates for decision-making, and multiple plausible future scenarios for risk assessment; ultra-rare classes with only a few examples remain challenging and should be interpreted with their support counts.}
\end{abstract}


\section{Introduction}
\IEEEPARstart{P}{hotovoltaic} (PV) solar power plants are now deployed at utility scale and have become an important pillar of the renewable energy portfolio \cite{sharadga2020time}. Maintaining high availability in these assets requires anticipating degradations and faults before they translate into energy losses or unsafe conditions. Prior work has addressed predictive maintenance and fault anticipation through data fusion, anomaly detection, and control-oriented strategies in energy and automation systems \cite{wei2021decisionfusion,maggipinto2022deepautoencoder,ahmed2019msthydropower,forouzanfar2025controlallocation}, and has leveraged degradation signals and high-dimensional sensor data for performance evaluation and process monitoring \cite{luo2019realtimeperformance,yan2014tensorprocessmonitoring}. Modern PV plants are heavily instrumented: supervisory control and data acquisition (SCADA) systems continuously record high-dimensional sensor streams (e.g., power, voltage/current, temperatures, weather, and curtailment status), while inverters and protection devices emit discrete fault/warning codes and maintenance logs. These two data modalities---dense multivariate time series and sparse marked event sequences---provide complementary views of the same physical system, yet they are rarely modeled jointly in a way that supports long-horizon simulation.

Generative modeling offers a natural route to such a ``digital twin.'' In natural language processing, large language models generate coherent futures by repeatedly predicting the next token; in multi-modal settings, vision--language models extend the same idea by learning shared token spaces and cross-attention mechanisms that couple heterogeneous modalities \cite{vaswani2017attention,lu2019vilbert}. A similar principle is attractive for PV systems: if we can learn a model that generates the next segment of sensor trajectories and the next fault/maintenance event conditioned on history, then we can roll the model forward to simulate plausible future operating scenarios and evaluate maintenance policies or dispatch strategies offline.

From an operations perspective, the PV forecasting problem is inherently dual-modal. On the continuous side, time-series generation aims to forecast future sensor trajectories and energy output under changing irradiance and operating regimes \cite{sharadga2020time}. On the discrete side, event generation aims to predict \emph{which} fault may occur and \emph{when}, supporting prognostics and predictive maintenance. These modalities are strongly coupled: rare inverter faults can trigger abrupt changes in electrical signals, while subtle trends in temperatures, currents, or communication status can precede the fault codes that eventually appear. Capturing these inter- and intra-modal dependencies is essential for reliable prediction and for simulation-based decision making.

Prior work typically focuses on one modality or couples them in limited ways. For time series, recurrent networks and transformer variants have been widely used, and recent work such as iTransformer improves long-horizon multivariate forecasting by inverting the tokenization to model sensor-wise dependencies efficiently \cite{liu2023itransformer}. For event sequences, temporal point processes (e.g., Hawkes processes) model event intensities \cite{laub2015hawkes}, while neural point processes and Transformer Hawkes Process (THP) use recurrent or attention-based representations to capture richer historical effects \cite{du2016recurrent,zuo2020transformer}. Score-matching-based approach for THP (SMURF-THP) \cite{li2023smurf} provides fixed-noise-level denoising score matching for event timing. Joint models have been explored (e.g., attentional twin RNNs and generative RNN-ODEs) \cite{xiao2017joint,li2022learning}, but they often struggle with the combination of high-dimensional long-range time series, sparse and highly imbalanced fault types, and the need for probabilistic long-horizon generation.

These considerations lead to three practical challenges highlighted in our PV setting. (i) \textbf{Rare events and long-term dependency}: critical failure modes may be infrequent, requiring models that can leverage long histories without losing signal. (ii) \textbf{Multi-scale time-series dynamics}: sensor streams exhibit both fast fluctuations and slower regime changes, making accurate long-range forecasting difficult. (iii) \textbf{Cross-modal representation}: the model must define suitable ``tokens'' for continuous and discrete data, fuse them without requiring perfect alignment, and produce probabilistic predictions for both trajectories and event time/type \cite{hullermeier2021aleatoric,kendall2017uncertainties}.

To address these challenges, we propose a multimodal generative framework that jointly models PV sensor time series and fault event sequences in a unified encoder--decoder architecture. Our encoder combines an iTransformer for time-series features with a THP encoder for marked events, and uses a co-attention module to explicitly model cross-modal interactions. From the fused latent representation, we perform conditional generation with generative decoders: a Conditional Latent Diffusion Model (CLDM) \cite{su2025diffusion} for multivariate sensor trajectory generation and a SMURF-THP for probabilistic event prediction \cite{li2023smurf}. The diffusion decoder operates in a learned latent space, enabling expressive distribution modeling while providing uncertainty estimates through stochastic sampling. Similarly, SMURF-THP leverages score matching to model the event arrival process, allowing probabilistic prediction of both event timing and type with quantified uncertainty. \jiayurev{Together, these decoders define a forward-in-time scenario generator: given an observed context $\mathcal{C}_t$, the model samples from $p_\Theta(\mathbf{X}_{t:t+H_{N}},\mathcal{E}_{t:t+H_{N}}\mid\mathcal{C}_t)$ over a specified number of rollout iterations $N_{\mathrm{iter}}$, where $H_N=\sum_{r=1}^{N_{\mathrm{iter}}}\Delta_r$ and $\Delta_r=\min(\hat{\tau}_r,7\ \mathrm{days})$ is the generated segment length at iteration $r$. Thus, the realized calendar span is induced by the iteration sequence and the seven-day per-iteration cap, and the resulting samples are used to summarize uncertainty rather than to claim deterministic extrapolation over the full observational record.}

Our contributions can be summarized as follows.
\begin{itemize}
    \item We formulate PV predictive maintenance as a \emph{multi-modal generative modeling} problem and propose a conditional generation framework that jointly simulates high-dimensional sensor time series and sparse fault event sequences.
    \item We design a unified encoder that integrates iTransformer-based time-series representations and Transformer Hawkes Process event representations through a co-attention fusion mechanism, enabling cross-modal dependency modeling without strict temporal alignment.
    \item We integrate CLDM sensor generation with SMURF-THP event prediction \cite{li2023smurf} for uncertainty-aware continuous and discrete outcomes.
    \item \jiayurev{On Red Rock Solar Site data, we show improved rare-event prediction and finite-iteration multimodal rollouts for maintenance planning.}
\end{itemize}

The rest of this paper is organized as follows. Section \ref{sec:related_work} discusses related work on event sequence modeling and time series forecasting. Section \ref{sec:proposed_method} details our proposed approach, including the model architecture, encoders, decoders, and conditional generation design. Section \ref{sec:exp} presents experimental results and Section \ref{sec:conclusion} concludes.

\section{Related Work}\label{sec:related_work}

Utility-scale photovoltaic (PV) plants generate \emph{continuous} SCADA/IoT streams (e.g., power, currents/voltages, inverter telemetry, and meteorological variables) alongside \emph{discrete} operational logs (e.g., alarms, fault codes, and maintenance work orders). 
Accurate modeling of both modalities is critical for forecasting, condition monitoring, and reliability-aware decision making.
Prior work spans PV power forecasting, PV fault diagnosis/predictive maintenance, temporal point processes for event prediction, and probabilistic generative modeling. 
However, most existing approaches either focus on forecasting the continuous dynamics while treating events as exogenous labels, or model event sequences without explicitly generating the associated multivariate sensor trajectories. 
This section reviews the most relevant literature and positions our multimodal generative simulator within these research threads.

\subsection{Time Series Prediction for Solar PV Systems}
PV power forecasting has been studied extensively across horizons ranging from seconds-ahead nowcasting to day-ahead and longer forecasts. 
Comprehensive reviews summarize the evolution from physical/parametric models and classical statistical approaches to modern machine learning and deep learning methods \cite{antonanzas2016pvforecasting}. 
Early data-driven work often relied on shallow models such as multilayer perceptrons \cite{popescu2009multilayer} and recurrent neural networks (RNNs) \cite{mellit2009recurrent}, later progressing to gated RNN variants for improved temporal credit assignment.
Deep architectures that combine representation learning with sequence modeling (e.g., autoencoders and LSTMs) were shown to be effective for PV output forecasting across multiple plants \cite{gensler2016solarforecast}. At the system level, multi-source fusion for solar plant power forecasting \cite{zhang2024dbnsolar} and uncertainty characterization for net load in renewable power systems \cite{yang2025netloaduncertainty} have been studied. 

More recently, attention mechanisms \cite{vaswani2017attention} and Transformer-style encoders have become prominent in time series forecasting, motivated by their ability to model long-range dependencies and complex cross-variable interactions.
Beyond generic Transformers, several architectures were proposed specifically to address long-horizon forecasting and long-context efficiency, including Informer \cite{zhou2021informer} and Autoformer \cite{wu2021autoformer}. 
Patching and channel-independent designs further improved scalability and performance on multivariate forecasting, as exemplified by PatchTST \cite{nie2023patchtst}. 
Task-general backbones such as TimesNet \cite{wu2023timesnet} extend these ideas by explicitly modeling multi-periodicity. 
In the PV domain, such models are typically trained on multivariate SCADA streams augmented with meteorological variables or numerical weather prediction features \cite{antonanzas2016pvforecasting}. 
Diffusion-based surrogate modeling and multi-fidelity calibration have also been explored  \cite{yan2025diffusionsurrogate}. 
Recent work on universal generative multimodal time series forecasting \cite{wu2026aurora} and multimodal extensions of inverted Transformers for time series \cite{mou2025mmitransformer} further illustrate the trend toward joint representation of multiple modalities. 
Our work builds on this forecasting lineage, but targets a \emph{generative} objective: producing realistic future sensor trajectories conditioned on historical context, which is essential for simulation and risk-sensitive downstream decision making.

\subsection{PV Fault Diagnosis, Predictive Maintenance, and Digital Twins}
In addition to forecasting, PV plants require robust condition monitoring to detect incipient failures and schedule maintenance.
Recent surveys highlight the rapid adoption of AI for PV predictive maintenance, including anomaly detection, fault classification, and remaining useful life estimation from operational data \cite{vichare2025pdmreview,ledmaoui2025predictive}. 
Distributed IoT and cloud-based frameworks have been proposed for predictive maintenance of solar panels \cite{mohammed2025iot}. 
Related work also covers PV layout and manufacturing inspection, wind-turbine fault diagnosis, and multisource fusion/prognostics \cite{mignoni2025layoutpv,zhao2025dbanpv,zhang2024windfault,wei2021decisionfusion,li2024selfsupervisedfusion,zhang2021multisensorprognostics}. 
Parallel to these efforts, digital twin (DT) concepts have been adapted to PV systems to support continuous synchronization between a physical plant and a data-driven or hybrid virtual model \cite{alao2024dtpvreview,song2023digitaltwin}. 
Most PV DT work emphasizes improved power prediction and adaptive model updating, including Transformer-based DT formulations for PV power prediction \cite{zhao2024dttransformer}. 
Despite this progress, PV DTs are commonly used as \emph{predictive} tools (estimating power or health indicators) rather than as \emph{stochastic simulators} capable of jointly generating plausible trajectories and discrete fault/maintenance event sequences over long horizons.
Our goal complements DT forecasting by providing a generative simulator for stress-testing, what-if analysis, and data-driven evaluation of maintenance policies.

\subsection{Event Sequence Prediction and Temporal Point Processes}
Discrete PV plant events—such as inverter trips, alarm bursts, or corrective maintenance actions—are naturally modeled as event sequences in continuous time.
Temporal point processes (TPPs), including Hawkes processes \cite{laub2015hawkes}, provide a principled framework for capturing self-excitation, inhibition, and cross-type triggering effects among events. 
Neural extensions replace hand-crafted kernels with sequence models, enabling flexible, history-dependent intensities.
Representative approaches include the recurrent marked temporal point process (RMTPP) \cite{du2016recurrent} and continuous-time LSTM formulations such as the Neural Hawkes Process \cite{mei2017neuralhawkes}. 
Transformer parameterizations further improve the modeling of long-range dependencies and inter-event interactions via self-attention \cite{zuo2020transformer}; extensions include sparse THP for long event sequences \cite{li2023sthp} and rotary position embedding-based variants \cite{gao2024rothp}. Recent surveys summarize modeling choices and evaluation protocols across the neural TPP literature \cite{shchur2021tppreview}.

A complementary line of work avoids explicit intensity modeling by directly parameterizing inter-event time distributions, enabling alternative likelihood formulations and more flexible sampling \cite{shchur2020intensityfree}. 
Motivated by limitations of autoregressive long-horizon sampling, recent TPP work explores parallel sequence diffusion and related generative formulations \cite{ludke2023addthin,zeng2024interacting,mukherjee2025adiff4tpp,zhou2023dltpp,kerrigan2026eventflow}, which is relevant when rare PV faults make rollouts difficult.

\subsection{Probabilistic Generative Modeling for Time Series and Events}
For simulation and decision support, it is often insufficient to provide point forecasts; instead, models must represent the conditional distribution over futures.
Classical probabilistic forecasting methods in energy systems include Bayesian and ensemble approaches, while modern deep generative models offer flexible distribution modeling and sampling.
For continuous time series, generative adversarial networks (GANs) were adapted to produce realistic synthetic sequences (e.g., TimeGAN) \cite{yoon2019timegan}, and have been used for model-free renewable scenario generation \cite{chen2018model}. 
Solar applications include synthetic PV/irradiance generation and scenario-generation surveys \cite{rosa2021unsupervised,lajic2024generative,losada2020soda,georgilakis2023review,turowski2024generating}. Diffusion/score-based models are a strong alternative \cite{ho2020denoising,song2021scorebased}, with time-series forecasting/imputation variants and surveys \cite{rasul2021autoregressive,tashiro2021csdi,shen2023timediff,kollovieh2023tsdiff,su2025diffusionsurvey}. For events, SMURF-THP~\cite{li2023smurf} and TPP diffusion methods such as ADD-THIN~\cite{ludke2023addthin} support non-autoregressive timing generation.

Our approach leverages these generative modeling advances to build a \emph{joint} multi-modal simulator: a CLDM model generates multivariate sensor trajectories with quantified uncertainty, while a SMURF-THP decoder generates fault events with associated timing and type. Both decoders are conditioned on a shared cross-modal latent representation, explicitly capturing the coupled dynamics between PV system states and operational events. 
This unified design enables coherent scenario generation where sensor trajectories and fault occurrences are jointly sampled, supporting risk-aware maintenance planning through probabilistic forecasts of both continuous system behavior and discrete fault processes.

\subsection{Joint Modeling of Event and Time Series Data}
A growing literature studies learning from datasets that contain both irregular event streams and regularly sampled time series. 
In automation applications, multisource fusion is a recurring theme for quality monitoring and predictive maintenance, including decision-level fusion and self-supervised heterogeneous fusion approaches reported  \cite{wei2021decisionfusion,li2024selfsupervisedfusion}. 
Early joint models couple RNN intensities with continuous observations (e.g., twin RNNs \cite{xiao2017joint}); related RPPN-style and RNN-ODE methods jointly forecast events and series \cite{xiao2017modeling,xiao2019learning,li2022learning}, and inverted Transformers handle multimodal series \cite{mou2025mmitransformer}. Continuous-time latents (e.g., latent ODEs \cite{rubanova2019latentode} and neural CDEs \cite{kidger2020neuralcde}) similarly support irregular sampling.

Despite these advances, joint models in the PV context typically remain \emph{predictive}: they aim to improve next-step forecasting or event prediction accuracy.
In contrast, maintenance planning and reliability analysis require \emph{rollout-capable} simulators that can generate coupled sensor trajectories and event sequences over long horizons, while preserving variability and avoiding the error accumulation common in purely autoregressive pipelines.
Our work addresses this gap by combining (i) Transformer-based encoders for each modality, (ii) co-attention fusion for cross-modal dependencies, and (iii) dual generative decoders-a latent diffusion model for probabilistic sensor trajectory generation and a SMURF-THP for event prediction \cite{li2023smurf}-enabling parallel, stochastic generation of both continuous trajectories and discrete events. Table~\ref{supp:tab:temporal-gen-comparison} summarizes representative methods across event diffusion/flow, joint event--time-series co-prediction, and time-series forecasting, and positions our framework in this landscape.

\section{Proposed Method}\label{sec:proposed_method}

\begin{figure*}[h]
  \centering
\includegraphics[width=0.78\textwidth]{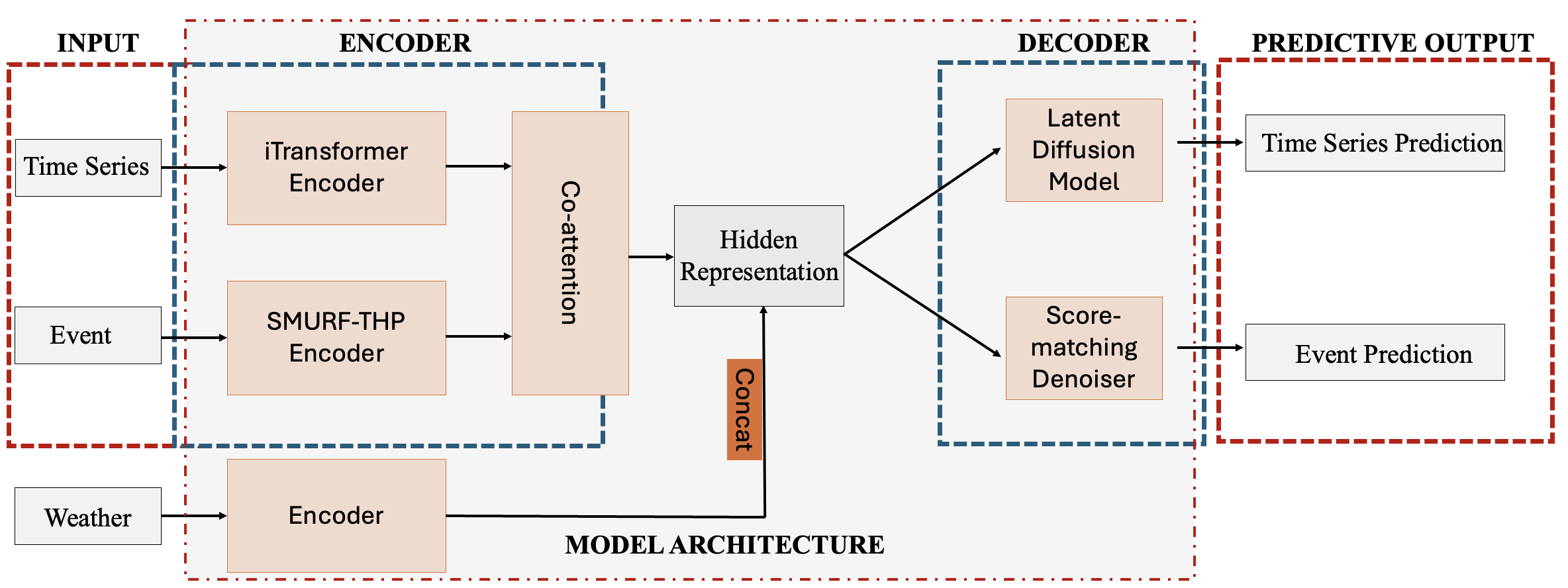}
  \caption{Overall encoder--decoder architecture: iTransformer and THP fused by co-attention; CLDM and SMURF-THP decoders.}
  \label{fig:architecture}
\end{figure*}

The proposed multi-modal generative model architecture follows an encoder-decoder structure designed for the integration of multi-modal data. As illustrated in Figure \ref{fig:architecture}, the architecture is divided into two main components: an encoder and a decoder. The encoder itself consists of two specialized modules: the 'iTransformer encoder \cite{liu2023itransformer}' for processing long time series data and the 'THP encoder \cite{zuo2020transformer}' for event sequences with a single mark: event type $k_i$ at timestamp $t_i$ (duration is excluded; see Section~\ref{sec:exp}). These modules extract and exchange critical features, ultimately fusing the latent representations by a 'Co-attention layer' for subsequent processing. Based on this fused context, the decoding stage employs two distinct generative mechanisms:

\begin{itemize}
    \item \textbf{Time Series Generation}: A \textbf{Conditional Latent Diffusion Model (CLDM)} generates future sensor trajectories. Unlike deterministic MLP decoders, this probabilistic module iteratively denoises sensor states conditioned on $\mathbf{h}$, enabling uncertainty quantification.
    \item \textbf{Event Prediction}: A \textbf{SMURF-THP Decoder} \cite{li2023smurf} predicts the next event's timing and type. It employs denoising score matching to model the continuous inter-arrival time distribution and a categorical head for event types.
\end{itemize}

\subsection{iTransformer Encoder}
The iTransformer encoder \cite{liu2023itransformer} addresses sequential dependencies in time series data using an attention mechanism applied to tokenized input features. Unlike vanilla Transformers \cite{vaswani2017attention} that treat each time step as a token (making attention quadratic in the lookback length), iTransformer inverts tokenization by treating each variate as a token whose embedding summarizes the lookback window. Self-attention is then applied across variates to capture cross-sensor dependencies, enabling efficient long-context multivariate forecasting.
\jiayurev{Formally, for a lookback matrix $\mathbf{X}_{1:L}\in\mathbb{R}^{L\times C}$ with $C$ variables, iTransformer constructs one token per variable,
\begin{equation}
    \mathbf{u}_c = \mathrm{Embed}(\mathbf{X}_{1:L,c}), \qquad c=1,\ldots,C,
\end{equation}
where $\mathbf{X}_{1:L,c}$ is the historical trajectory of variable $c$. Multi-head self-attention is then applied over the token set $\{\mathbf{u}_c\}_{c=1}^{C}$ rather than over time-step tokens. This design is well matched to the PV dataset because the 61 synchronized variables span system-, inverter-, and module-level measurements, and fault anticipation often depends on cross-variable relationships such as interactions among power, voltage/current, temperature, weather, and status channels.}

The iTransformer approach is particularly effective for long multivariate lookbacks: treating each of the 61 synchronized PV variables as a token makes cross-sensor attention natural for fault anticipation. A schematic comparison with a vanilla time-step Transformer is given in Figure~\ref{supp:fig:itrans}.

\subsection{Transformer Hawkes Process Encoder}
We model each event by its timestamp $t_i$ and a single \emph{mark} $k_i$ (event type). Duration is not used as a mark; the marked point process is thus $(t_i, k_i)$ with $k_i \in \{1,\ldots,K\}$ the fault type. The THP encoder \cite{zuo2020transformer} models these event sequences by leveraging a Transformer-based architecture to parameterize the temporal dynamics of event time and type distributions. Unlike traditional Hawkes processes, which rely on predefined parametric intensity functions, THP integrates self-attention mechanisms to dynamically model event dependencies, capturing both long-range historical influences and short-term interactions.

THP encoder details appear in Section~\ref{supp:sec:thp_encoder_formulation}.

\subsection{Co-Attention Layer}
\label{sec:coattention}
The proposed co-attention module facilitates cross-modal learning and feature enhancement by dynamically fusing hidden representations from time series and event encoders. Adapted from the co-attention transformer architecture introduced in ViLBERT \cite{lu2019vilbert}, this multi-modal framework processes both temporal and event-based data through an interaction mechanism that captures interdependencies between the two modalities.

The module extends the standard transformer architecture by not only computing query $(Q)$, key $(K)$, and value $(V)$ matrices, but also incorporating cross-modal attention layers. Specifically, the keys and values derived from one modality are integrated into the multi-head attention blocks of the complementary modality, enabling effective information exchange. This cross-attention fusion captures latent correlations between temporal patterns and event sequences, ensuring that sensor trends inform event predictions and vice versa.

Figure~\ref{supp:fig:coattention} illustrates multimodal fusion. Event-attending-to-time-series co-attention is
\[
A_e = \mathrm{softmax}\!\left(\frac{Q_e K_{ts}^\top}{\sqrt{d_k}}\right)V_{ts},
\]
with event queries $Q_e$ and time-series keys/values $K_{ts},V_{ts}$. Symmetrically,
\[
A_{ts} = \mathrm{softmax}\!\left(\frac{Q_{ts} K_e^\top}{\sqrt{d_k}}\right)V_e,
\]
with time-series queries $Q_{ts}$ and event keys/values $K_e,V_e$ ($d_k$ is the key dimension). The outputs $A_e$ and $A_{ts}$ represent the time-series-informed event representation and the event-informed time-series representation, respectively. These bidirectional cross-attention operations---together with feed-forward layers and residual connections---dynamically fuse the two modalities, enhancing the model's ability to capture inter-dependencies between temporal patterns and discrete event sequences. By integrating co-attention in this manner, our work demonstrates a more comprehensive and accurate predictive framework for real-world applications.
\jiayurev{To keep notation consistent with Figure~\ref{supp:fig:coattention} in the supplementary material, we use $\mathcal{H}_t$ only for the event history set. The quantities $H_e^j$ and $H_{ts}^i$ shown in the figure denote layer-indexed hidden-state matrices from the event and time-series streams, respectively; they are not the same object as $\mathcal{H}_t$. When the layer index is not essential, we write these hidden-state matrices compactly as $\mathbf{H}_e$ and $\mathbf{H}_{ts}$, and we denote the fused latent context by $\mathbf{h}$. The attention matrices are linear projections of these hidden states. For event attending to time series, $Q_e=\mathbf{H}_eW_Q^e$, $K_{ts}=\mathbf{H}_{ts}W_K^{ts}$, and $V_{ts}=\mathbf{H}_{ts}W_V^{ts}$. For time series attending to events, $Q_{ts}=\mathbf{H}_{ts}W_Q^{ts}$, $K_e=\mathbf{H}_eW_K^e$, and $V_e=\mathbf{H}_eW_V^e$.}
\jiayurev{The final decoders use the fused context rather than the raw THP intensity parameters. Specifically, $\mathbf{H}_{e}=\mathrm{THPEncoder}(\mathcal{H}_{t})$ and $\mathbf{H}_{ts}=\mathrm{iTransformer}(\mathbf{X}_{1:L})$ are fused as $\mathbf{h}=\mathrm{Fuse}(\mathbf{H}_{e},\mathbf{H}_{ts})$. The SMURF-THP event-time network is parameterized as $s_\theta(\tilde{x},\mathbf{h})$, and the event-type head is parameterized as $p_\omega(k|\mathbf{h},x)$. Thus, the THP intensity parameters $(\alpha_k,\mathbf{w}_k,b_k)$ are distinct from the decoder parameters $(\theta,\omega)$; the former define the event-history representation and baseline formulation, whereas the latter are used for the proposed next-event decoder conditioned on $\mathbf{h}$.}

\subsection{Conditional Latent Diffusion for Time Series}
\label{sec:diffusion_ts}
To generate probabilistic future sensor trajectories, we implement a Conditional Latent Diffusion Model. This approach replaces traditional deterministic forecasters with a generative process that learns the conditional distribution of future states $p(\mathbf{x}_{pred} | \mathbf{h})$ while supervising latent alignment.

\paragraph{Diffusion Process}
We model future windows $\mathbf{x}_0 \in \mathbb{R}^{T \times C}$ ($T$ horizon, $C$ channels) via forward diffusion in latent space. The noisy state at step $t$ is
\begin{equation}
    q(\mathbf{z}_t|\mathbf{z}_0) = \mathcal{N}\!\left(\sqrt{\bar{\alpha}_t}\,\mathbf{z}_0,\,(1-\bar{\alpha}_t)\mathbf{I}\right),
\end{equation}
where $\mathbf{z}_0 = \text{CoAttention}(\mathbf{x}_0, E)$ is the latent embedding of the target future sequence, $E$ is the event-related input, and $\bar{\alpha}_t$ is determined by a cosine variance schedule (here $t$ indexes the diffusion step, not physical time). This ensures smooth noise transitions compared to linear schedules.

\paragraph{Conditional Reverse Process}
The generative process learns to invert this noise by predicting the added noise $\boldsymbol{\epsilon}_\theta(\mathbf{z}_t, t, \mathbf{h})$ conditioned on the multimodal system context $\mathbf{h}$. The denoising network employs a Transformer-based architecture that takes the noisy latent $\mathbf{z}_t$, diffusion step embedding $t$, and the context $\mathbf{h}$ to estimate the noise component. The estimated clean latent $\hat{\mathbf{z}}_0$ is then recovered from the noisy input and predicted noise:
\jiayurev{More specifically, the conditional denoiser is a mapping $\boldsymbol{\epsilon}_\theta:(\mathbf{z}_t,e(t),\mathbf{h})\mapsto\widehat{\boldsymbol{\epsilon}}$, where $e(t)$ denotes the sinusoidal diffusion-step embedding. The noisy latent $\mathbf{z}_t$ is first projected to the model dimension and combined with $e(t)$. The fused multimodal context $\mathbf{h}$ is injected through cross-attention and residual feed-forward blocks, so that the reverse process is conditioned on both sensor-history and event-history information. The final projection maps the hidden state back to the latent dimension to estimate the injected Gaussian noise. This follows the standard conditional diffusion parameterization, with conditioning provided by the co-attention representation rather than by class labels or external text prompts.}
\begin{equation}
    \hat{\mathbf{z}}_0 = \frac{\mathbf{z}_t - \sqrt{1-\bar{\alpha}_t}\,\boldsymbol{\epsilon}_\theta(\mathbf{z}_t, t, \mathbf{h})}{\sqrt{\bar{\alpha}_t}}.
\end{equation}
Finally, a projection head maps this recovered latent back to the sensor data space: $\hat{\mathbf{x}}_0 = f_{\text{proj}}(\hat{\mathbf{z}}_0)$.

\paragraph{Hybrid Training Objective}
To accelerate convergence and keep high-fidelity latents, we use a hybrid loss with diffusion noise reconstruction and projection supervision:
\begin{align}
    \mathcal{L}_{\text{diff}} &= \mathbb{E}_{t, \mathbf{z}_0, \boldsymbol{\epsilon}}\!\left[\|\boldsymbol{\epsilon} - \boldsymbol{\epsilon}_\theta(\mathbf{z}_t,\, t,\, \mathbf{h})\|^2\right] \\
    \mathcal{L}_{\text{proj}} &= \|\mathbf{x}_0 - f_{\text{proj}}(\hat{\mathbf{z}}_0)\|^2 \\
    \label{eq:lossts}
    \mathcal{L}_{\text{TS}} &= \mathcal{L}_{\text{diff}} + \lambda_{\text{proj}} \mathcal{L}_{\text{proj}}
\end{align}
where $\lambda_{\text{proj}}$ is a weighting factor. The projection loss $\mathcal{L}_{\text{proj}}$ directly enforces that the denoised latent $\hat{\mathbf{z}}_0$ decodes to the correct sensor values, acting as a strong regularization term that stabilizes the training of the generative model.
\jiayurev{Although $\mathcal{L}_{\text{diff}}$ has a squared-error form, it should not be interpreted as ordinary deterministic regression to the conditional mean. The target is the injected random noise under randomly sampled diffusion steps, and the learned denoiser parameterizes the conditional reverse process. Stochasticity is therefore introduced during generation by sampling the initial latent noise and applying the reverse denoising chain, while point summaries are computed only after sampling for evaluation.}
\jiayurev{The time-series decoder is modular: CLDM is used here because stochastic samples and uncertainty bands are useful for maintenance-risk analysis, but the same fused context $\mathbf{h}$ could condition deterministic forecasters when only point prediction is required.}
\jiayurev{The rationale for retaining a generative decoder is that PV operation is heteroscedastic and regime dependent: similar historical states may lead to different future trajectories because of weather variability, curtailment, inverter state, and latent degradation. A deterministic decoder can be appropriate for estimating the conditional mean, but it cannot directly represent multi-modal futures or low-probability high-impact deviations. The CLDM decoder is therefore used to model $p(\mathbf{x}_{0:T}\mid \mathbf{h})$ as a distribution, which is the quantity needed for risk-aware maintenance and scenario analysis.}

\paragraph{Inference and Uncertainty}
At inference, sampling $\mathbf{z}_T \sim \mathcal{N}(0, \mathbf{I})$ and denoising over $S$ steps yields $\hat{\mathbf{z}}_0$, then $\hat{\mathbf{x}}_0=f_{\mathrm{proj}}(\hat{\mathbf{z}}_0)$. Repeated draws give trajectories $\hat{\mathbf{x}}_0^{(k)}$ whose mean and std.\ summarize the forecast and uncertainty.

\subsection{SMURF-THP for Event Prediction}
For discrete fault/maintenance events we use denoising score matching with a THP backbone \cite{zuo2020transformer}. Marks are \emph{type only} ($(t_i,k_i)$; no duration). SMURF-THP~\cite{li2023smurf} applies fixed-noise-level score matching to next-event timing rather than a full DDPM. Unlike the multi-step CLDM used for continuous time-series generation, the event decoder does not approximate an entire diffusion trajectory with one coarse discretization step. Instead, it targets the next positive inter-event interval. To guarantee positive inter-event times $\tau > 0$, we work in log-space with $x = \log \tau$, perturb and denoise $x$, and recover $\tau = \exp(x)$. This formulation is computationally efficient while still learning the score of the conditional inter-event-time distribution given the multimodal latent $\mathbf{h}$.
\jiayurev{The event decoder  follows SMURF-THP and uses denoising score matching at a fixed noise level to learn the score of the conditional log-inter-event-time density for the next event. This avoids intensity integration while providing sampling-based uncertainty intervals. A full multi-step diffusion process over complete event sequences may better support long-horizon non-autoregressive event generation, but it is more computationally demanding and is left for future work.}

\paragraph{Target variable and positivity constraint}
Given history $\mathcal{H}_t=\{(t_i, k_i)\}_{i=1}^{N_t}$, the next interval is $\tau = t_{i+1} - t_i>0$. For positivity under Gaussian perturbation and Langevin sampling we set
\begin{equation}
    x = \log \tau \in \mathbb{R}, \qquad \tau = \exp(x).
\end{equation}
Perturbing and sampling $x$ avoids negative inter-event times that would arise from applying Gaussian noise directly to $\tau$.

\paragraph{Denoising score matching formulation}
The decoder learns $s_\phi(x|\mathbf{h}) \approx \nabla_x \log p(x|\mathbf{h})$ by denoising score matching. Unlike multi-step DDPM, SMURF-THP uses a fixed noise level on the next log inter-event interval. Training perturbs $x = \log \tau$ as
\begin{equation}
    \tilde{x} = x + \sigma \xi, \quad \xi \sim \mathcal{N}(0,1),
\end{equation}
where $\sigma$ is fixed (unlike DDPM's schedule). Denoising minimizes
\begin{equation}
    \mathcal{L}_{\text{SMURF-time}} =
    \tfrac{1}{2}
    \mathbb{E}_{x, \tilde{x}}
    \big[
        \|
            \psi_\theta(\mathbf{h}, \tilde{x})
            - 
            \tfrac{x - \tilde{x}}{\sigma^2}
        \|^2
    \big],
\end{equation}
where $\psi_\theta$ approximates the score $\nabla_{\tilde{x}} \log q(\tilde{x}|x)$ and encourages the learned score to move $\tilde{x}$ toward the true $x = \log\tau$. We recover the inter-event time as $\tau = \exp(x)$. The key distinction from full DDPM is that SMURF-THP learns a local conditional score for a one-dimensional next-event timing variable, rather than a multi-step reverse process over a complete event sequence. This keeps the event decoder lightweight while retaining the probabilistic sampling benefits of score-based generative modeling.

\paragraph{Event type prediction}
Event type $k_{i+1}$ uses a softmax on $\mathbf{h}$ and timing $x = \log\tau$:
\begin{equation}
    p_\omega(k_{i+1}|\mathbf{h}, x)
    = 
    \left[\mathrm{softmax}(g_\omega([\mathbf{h};x]))\right]_{k_{i+1}},
\end{equation}
with $x = \log \tau$, trained by cross-entropy
\begin{equation}
    \mathcal{L}_{\text{SMURF-type}} 
    = \mathbb{E}_{(\mathbf{h}, x, k)} [-\log p_\omega(k|\mathbf{h}, x)].
\end{equation}
\jiayurev{During stochastic rollout of the proposed model, the next event type is sampled as $k_{i+1}\sim p_\omega(k\mid\mathbf{h},x)$ to preserve event-type uncertainty. During deterministic evaluation, we use $\hat{k}_{i+1}=\arg\max_k p_\omega(k\mid\mathbf{h},x)$ so that F1-scores are computed from a single predicted label.}

\paragraph{Joint objective}
The full SMURF-THP decoder loss combines the score-matching and classification components:
\begin{equation}
    \mathcal{L}_{\text{SMURF}} = \mathcal{L}_{\text{SMURF-time}} + \lambda_{\text{type}}\mathcal{L}_{\text{SMURF-type}},
\end{equation}
where $\lambda_{\text{type}}$ balances the influence of temporal accuracy and type prediction. This formulation replaces the conventional negative log-likelihood used in Hawkes processes, allowing stable training even when event intervals are irregular or non-exponential.

\paragraph{Conditional generation}
At inference, given $\mathbf{h}$, the next log inter-event time is sampled by Langevin updates
\begin{equation}
    x^{(n+1)} = x^{(n)} + \tfrac{\epsilon}{2}s_\phi(x^{(n)}|\mathbf{h}) + \sqrt{\epsilon}\, \eta, \quad \eta \sim \mathcal{N}(0, 1),
\end{equation}
where $\epsilon$ is the step size. We recover the inter-event time as $\tau = \exp(x^{(n+1)})$ and the next timestamp as $t_{i+1} = t_i + \tau$. The corresponding event type $k_{i+1}$ is then drawn from the learned categorical distribution $p_\omega(k|\mathbf{h}, x^{(n+1)})$. Repeating this process produces an entire event sequence consistent with the latent co-attention context, with $\tau > 0$ guaranteed by the log-space parameterization.

\paragraph{Interpretation and benefits}
By framing temporal event prediction as a denoising score matching task, SMURF-THP eliminates the need for explicit intensity integration and supports uncertainty-aware predictions. \jiayurev{The event decoder should be interpreted as single-noise-level denoising score matching for the next log inter-event interval, not as a full multi-step DDPM over an entire event sequence.} This design choice balances computational efficiency with the benefits of score-based generative modeling. The combination of denoising score matching for time prediction and cross-entropy for type classification allows the model to flexibly capture multimodal event distributions and stochastic inter-event intervals. When jointly trained with the DDPM time-series decoder, both modules share the same latent context $\mathbf{h}$, aligning continuous sensor dynamics with discrete fault occurrences for fully coupled system generation. While the current fixed-noise-level score-matching approach provides a lightweight solution, future extensions could incorporate a full multi-step DDPM framework for event prediction, enabling more sophisticated noise scheduling and iterative refinement of event timing predictions.
\jiayurev{This choice is also appropriate for sparse PV fault logs. Direct likelihood training for neural point processes requires integrating type-specific intensities over irregular intervals, which can be sensitive when events are sparse and inter-arrival times vary widely. Learning the score of $x=\log\tau$ avoids a positivity constraint on $\tau$, supports sampling-based uncertainty intervals.}

\subsection{Joint Objective and Training}
The joint training objective is
\begin{equation}
    \mathcal{L}_{\text{total}} 
    = 
    \mathcal{L}_{\text{TS}}
    + 
    \lambda_{\text{SMURF}}
    \mathcal{L}_{\text{SMURF}},
\end{equation}
where $\mathcal{L}_{\text{TS}} = \mathcal{L}_{\text{diff}} + \lambda_{\text{proj}} \mathcal{L}_{\text{proj}}$ is the diffusion-based time series loss introduced in Equation~\ref{eq:lossts}, and $\lambda_{\text{SMURF}}$ balances the contributions of event modeling and time-series forecasting. Both decoders share the latent co-attention representation $\mathbf{h}$, ensuring that the temporal dependencies across sensors and events are jointly exploited.
\jiayurev{Loss weights and architecture are chosen on the chronological train/validation split only (test unused for tuning). Reported runs use
\[
\mathcal{L}_{\mathrm{total}} = 0.7\bigl(\mathcal{L}_{\mathrm{diff}} + 1.5\,\mathcal{L}_{\mathrm{proj}}\bigr) + 2.0\,\mathcal{L}_{\mathrm{event}} + 0.1\,\mathcal{L}_{\mathrm{time}},
\]
with $\mathcal{L}_{\mathrm{event}}$/$\mathcal{L}_{\mathrm{time}}$ the SMURF-THP type/time terms. Splits: train 2019-05-10--2020-05-08, val 2020-05-09--2020-07-20, test 2020-07-21--2021-05-09; see Table~\ref{supp:tab:reproducibility_settings}.}

\subsection{\jiayurev{Finite-Step Rollout Generation}}
To simulate operational behavior, we use a \jiayurev{strictly chronological, history-conditioned rollout with a specified number of generation iterations $N_{\mathrm{iter}}$. At each iteration, the model conditions only on past observations and previously generated outputs, uses the same seven-day operational prediction window as the event-evaluation protocol, and generates a sensor segment capped at seven days. If the predicted next event falls within this window, the event is appended to the generated history; otherwise, the rollout advances by the generated seven-day segment without recording an event for that iteration. Figure~\ref{supp:fig:generation_method} shows the rollout structure, and Figure~\ref{supp:fig:generation_flowchart} provides the corresponding step-by-step workflow.}
\section{Experiments and Results}\label{sec:exp}

\subsection{Data Description and Preparation}

The data set consists of time series data from various levels of the solar energy system and real-time fault data to enrich the training samples.

\subsubsection{15-Minute Interval Time Series}
The dataset comprises 61 variables at different operational levels of the photovoltaic plant, as described in Table \ref{t:variables}. These variables capture the dynamics of the granular system at a resolution of 15 minutes, enabling holistic performance monitoring.

\begin{table}[ht]
\centering
\caption{Variables at Different Operational Levels of the PV Plant}
\label{t:variables}
\setlength{\tabcolsep}{4pt}
\begin{tabular}{l c r}
\toprule
  \textbf{Data Level} & \textbf{No. of Variables} & \textbf{Examples} \\
\midrule
  System Level & 47 & curtail, weather, feeder \\
  Inverter Level & 11 & power output (current, voltage) \\
   Module Level & 3 & status, T-IGBT \\
\bottomrule
\end{tabular}
\end{table}

\begin{itemize}
  \item \textbf{System Level:} The dataset includes 47 variables at the system level, covering both operational metrics and weather-related parameters, as well as data from three feeders.
  \item \textbf{Inverter Level:} At the inverter level, 11 variables are recorded, with a focus on the power output, represented by voltage (V), current (I), and direct current (DC) measurements.
  \item \textbf{Inverter Module Level:} Data at the inverter module level consists of 3 key variables (excluding the fault data), which include system status (warning codes), cabinet temperature, and T-IGBT (Insulated Gate Bipolar Transistor temperature).
\end{itemize}

The processed data is organized into a multi-dimensional array with shape $(200, 61, 70176)$, where 200 represents the number of inverter modules, 61 denotes the total number of variables across all levels, and 70,176 corresponds to the number of time-stamped data points collected at 15-minute intervals. At this resolution, 70,176 timestamps span $70{,}176 \times 15\,\text{min} \approx 731$ days (almost two years).

We use a time-based split into training, validation, and test sets so that evaluation reflects forecasting on genuinely future, unseen periods: the training and validation portions use the earliest contiguous portion of the timeline, and the test set follows them. This policy avoids look-ahead bias and is strongly preferred for forecasting and predictive maintenance. 
\jiayurev{Each sample is constructed from a three-month historical input window followed by a seven-day prediction window. Let $s_i$ denote the input-window start time, $o_i=s_i+90$ days the forecast origin, and $e_i=o_i+7$ days the end of the prediction window. Chronological assignment is performed by $(o_i,e_i)$ rather than by $s_i$: training samples satisfy $e_i\leq$ the train/validation boundary, validation samples are assigned between the validation and test boundaries, and test samples have $o_i$ on or after the test boundary. Boundary-crossing samples are discarded. Therefore, for the first test prediction on 2020-7-21, the three-month lookback may contain observations before the test start date, but these observations are historical context available before the forecast origin, not future targets.}
\jiayurev{This protocol avoids look-ahead leakage because any sample is assigned according to the time interval being predicted. Thus, a test input window may overlap with the earlier calendar period as historical context, but no test target, next-event label, normalization statistic, or hyperparameter-selection information is taken from held-out test timestamps. A rolling-origin validation protocol would provide an additional robustness analysis across multiple forecast origins; however, it is not required to establish the absence of temporal leakage. Our chronological holdout split follows the standard forecasting convention while preserving a sufficiently long contiguous test period for rare-event evaluation.}

\subsubsection{Real-Time Fault Data}

Real-time fault data is captured at the inverter module level. In this work, we include the 12 most common fault types. Table~\ref{supp:t:fault_descriptions} describes these fault codes.

The real time fault data undergoes a preprocessing pipeline to ensure consistency and suitability for event modeling. Specifically, our processing approach applies the following rules:
\begin{itemize}
  \item The consecutive rows with the same fault code are aggregated as a single event.
  \item If fewer than three rows indicating a 'normal' state (fault code 0) occur between instances of the same fault code, these rows are ignored, and the event is extended.
  \item Events that last less than 10 minutes are discarded to filter out transient noise.
\end{itemize}

These preprocessing steps standardize the fault data, enabling its effective use as input for event sequence modeling. Table \ref{supp:t:real_time_fault_data} presents an example of the processed data with a sequence length of 6, where each entry records the start time, gap time between consecutive events, event type, and the duration, respectively. For event modeling in this paper, we use only \emph{event time} (or equivalently gap time) and \emph{event type} as the mark; duration is recorded in the table for reference but is \emph{not} used as a mark in the THP/SMURF-THP model.

\subsection{Model Performance Evaluation}

\subsubsection{Time Series Prediction Performance}
\begin{figure}[htbp]
  \centering
  \includegraphics[width=0.95\columnwidth]{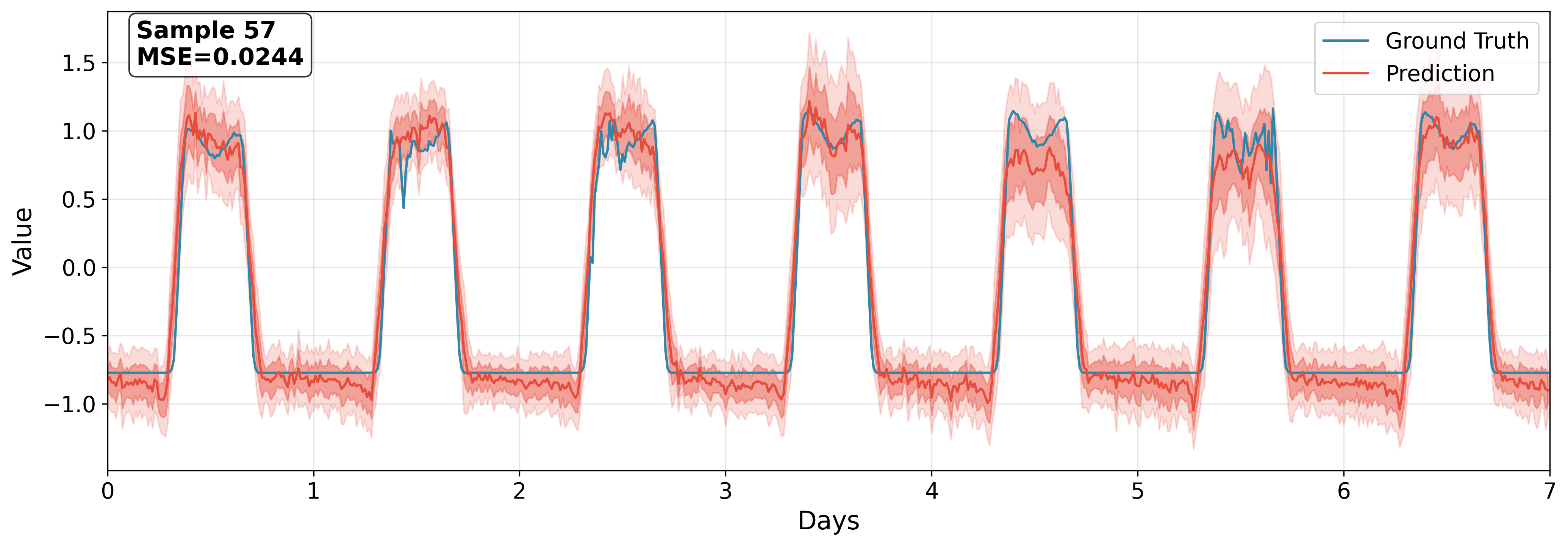}
  \caption{Power-signal forecast on a representative test sample (more samples in Supp.~Section~\ref{supp:app:more_power_vis}).}
  \label{fig:ts_and_next_event}
\end{figure}

Figure~\ref{fig:ts_and_next_event} presents the time series generation results over a seven-day span on a representative test sample, comparing the generated power signal to the ground truth. The CLDM prediction (red line) closely tracks the true signal (blue line), and the red shaded regions depict confidence intervals that effectively envelop the ground-truth trajectory. The varying band width reflects dynamic uncertainty estimation over the daily generation cycle. \jiayurev{Additional qualitative results on six test samples with per-sample MSE are in Section~\ref{supp:app:more_power_vis} (Figure~\ref{supp:fig:more_power_visualization_mse}).}

\jiayurev{Predictive uncertainty is state-dependent (higher during morning irradiance ramps than at night); see Figure~\ref{supp:fig:uncertainty_dynamics}. Using 20 DDPM samples with 50 diffusion steps, CRPS is 0.2844, with empirical coverages 76.66\% and 81.96\% for nominal 90\% and 95\% intervals (useful but under-calibrated).}

\subsubsection{Event Type Prediction Performance}

\paragraph{Event Prediction Protocol}
Event evaluation is \textbf{next-event prediction} in a \textbf{rolling} setting: at each step we condition on past events and the corresponding time-series history, and predict the next event type and next event time (inter-event interval $\tau$). \textbf{Input context:} we use a fixed number of past events (context length 10) and a matching 3-month time-series lookback window. Evaluation is \textbf{strictly next-event}: at test time we do not teacher-force; we feed the model the ground-truth history up to time $t$, then predict the single next event type $\hat{k}$ and time $\hat{t}$. \jiayurev{The event label is defined within a seven-day operational prediction window. This window is chosen to match short-term maintenance planning and because events much farther in the future may not have sufficiently strong or actionable precursors at the current forecast origin.} \textbf{Time error:} we compute MAE in days between predicted and ground-truth next event time on the test set over all next-event steps. \textbf{Rare classes:} Given that all event types in the dataset are relatively infrequent, we use Macro F1-score to provide an evaluation that accounts for the inherent rarity across the entire label distribution. \jiayurev{Support counts refer to labeled next-event samples whose next event occurs within the seven-day prediction window; events farther than the prediction window are not included as labels for that forecast origin. Encoded class 2 is omitted from class-level reporting because it has no test examples in the seven-day prediction-window labels. We report support in the ablation table so that low F1 values for ultra-rare classes are not over-interpreted. For example, type 42 has only 4 validation examples and 5 test examples, making its per-class F1 highly variable.}
\jiayurev{Class imbalance is handled using training-split information only. Let $n_k^{\mathrm{tr}}$ denote the number of training next-event labels of type $k$ and let $\mathcal{K}_{\mathrm{tr}}=\{k:n_k^{\mathrm{tr}}>0\}$. We use inverse-square-root type weights

\begin{equation}
\begin{aligned}
\omega_k
&=
\frac{
\sqrt{
\displaystyle
\frac{\sum_{j\in\mathcal{K}_{\mathrm{tr}}} n_j^{\mathrm{tr}}}
{|\mathcal{K}_{\mathrm{tr}}|\,n_k^{\mathrm{tr}}}
}
}{
\displaystyle
\frac{1}{|\mathcal{K}_{\mathrm{tr}}|}
\sum_{\ell\in\mathcal{K}_{\mathrm{tr}}}
\sqrt{
\frac{\sum_{j\in\mathcal{K}_{\mathrm{tr}}} n_j^{\mathrm{tr}}}
{|\mathcal{K}_{\mathrm{tr}}|\,n_\ell^{\mathrm{tr}}}
}
}.
\end{aligned}
\end{equation}
so that the average weight over observed training classes is one. The event-type loss is a weighted focal cross-entropy,
\begin{equation}
    \mathcal{L}_{\mathrm{type}}
    = -\omega_{y}(1-p_{y})^{\gamma}\log p_{y},
\end{equation}
with $\gamma=1.0$, where $p_y$ is the predicted probability of the true event type. In addition, the training loader uses an event-balanced sampler with sample weight proportional to $(n_{y_i}^{\mathrm{tr}})^{-\alpha}$, $\alpha=0.5$, capped at a maximum weight of 10. Validation and test labels are not used to compute these weights or sampling probabilities.}

\paragraph{Event Prediction Performance Comparison}
\jiayurev{Table~\ref{tab:f1_comparison_three_models} summarizes the final event-type F1 results on the held-out test set. The proposed model obtains the highest aggregate event-type performance, with Macro F1 of 0.3323 and Weighted F1 of 0.3565. It improves several classes with nonzero test support, including types 18, 19, 36, 55, 86, 95, and 100.}
\begin{table}[ht]
    \centering
    \caption{\jiayurev{Per-type next-event F1: baselines vs.\ proposed model.}}
    \label{tab:f1_comparison_three_models}

    \jiayurev{
    \begin{tabular}{cccc}
        \toprule
        \multirow{2}{*}{Fault Type}
        & \multicolumn{2}{c}{Baseline Models}
        & \multirow{2}{*}{\textbf{Proposed Model}} \\
        \cmidrule(lr){2-3}
        & RNNPP & Event-only & \\
        \midrule
        4   & 0.0000          & 0.0000          & 0.0000          \\
        18  & 0.5000          & 0.5000          & \textbf{0.5556} \\
        19  & 0.0930          & 0.1138          & \textbf{0.2017} \\
        36  & 0.1636          & 0.1250          & \textbf{0.1839} \\
        40  & \textbf{0.5766} & 0.2500          & 0.5326          \\
        42  & 0.0000          & \textbf{0.3636} & 0.1667          \\
        53  & \textbf{0.2080} & 0.0000          & 0.1594          \\
        55  & 0.4714          & 0.4662          & \textbf{0.4857} \\
        67  & \textbf{0.4231} & 0.2876          & 0.4091          \\
        86  & 0.3750          & 0.4762          & \textbf{0.5455} \\
        95  & 0.2941          & 0.4000          & \textbf{0.4516} \\
        100 & 0.2051          & 0.1695          & \textbf{0.2958} \\
        \midrule
        Micro    & 0.3185 & 0.2209 & \textbf{0.3390} \\
        Macro    & 0.2758 & 0.2627 & \textbf{0.3323} \\
        Weighted & 0.3350 & 0.2299 & \textbf{0.3565} \\
        \bottomrule
    \end{tabular}
    }
\end{table}

The proposed model integrates dense time series features and event sequence data via a co-attention mechanism, enabling a more nuanced understanding of event interdependencies. \jiayurev{Overall, the proposed model achieves the best aggregate event-type performance in the final test evaluation, with Macro F1 of 0.3323 and Weighted F1 of 0.3565. However, the class-wise results remain heterogeneous, reflecting the imbalanced and sparse nature of marked PV fault events. The proposed model improves several event types whose precursors benefit from multimodal sensor-event conditioning, while event-only or point-process baselines can remain competitive for classes driven primarily by recurrence structure or with very small support.}
\jiayurev{ The results indicate that cross-modal conditioning improves several low-frequency or moderate-frequency classes with enough statistical support in the seven-day prediction-window labels, including types 18, 19, 36, 55, 86, 95, and 100 in the test split, while classes with only a few labeled occurrences remain statistically difficult. Type 42, for example, has 4 validation examples and 5 test examples, and the event-only baseline performs better for this type. For such ultra-rare classes,  a purely data-driven classifier has limited evidence from which to estimate class-specific decision boundaries.}

\begin{figure}[t]
    \centering
    \includegraphics[width=0.95\columnwidth]{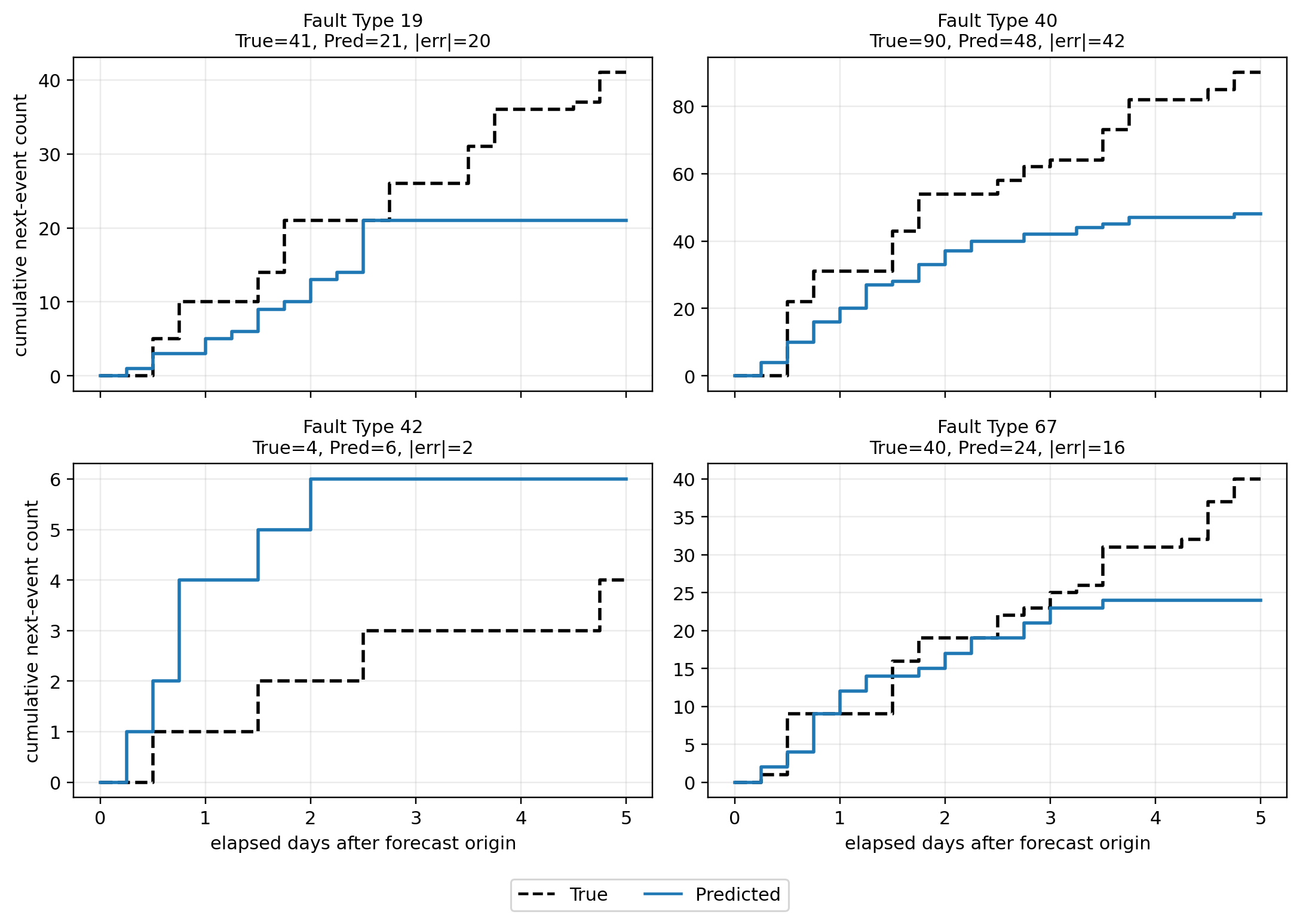}
    \caption{\jiayurev{Five-day origin-aggregated next-event counting processes (true dashed, predicted solid) for types 19, 40, 42, and 67.}}
    \label{fig:counting_process_selected_types}
\end{figure}

\jiayurev{Dependencies among fault types are handled implicitly by event-type embeddings and masked self-attention over the marked event history. Because the keys and values contain previous event types and times, the event encoder can represent cross-type recurrence patterns before the co-attention layer further modulates them with sensor context. We also evaluate origin-aggregated next-event counting processes. For test origin $i$ with true/predicted next event $(\tau_i,k_i)$ and $(\hat{\tau}_i,\hat{k}_i)$, and elapsed $u\in[0,5]$ days,
\[
N_k^{\mathrm{true}}(u)=\sum_i \mathbf{1}(\tau_i\le u,\ k_i=k),\qquad
N_k^{\mathrm{pred}}(u)=\sum_i \mathbf{1}(\hat{\tau}_i\le u,\ \hat{k}_i=k).
\]
Figure~\ref{fig:counting_process_selected_types} compares the true and predicted cumulative counts for fault types 19, 40, 42, and 67. These types cover relatively frequent fault categories, including types 19, 40, and 67, and a sparse fault category, type 42.  The curves show that the model captures part of the early accumulation pattern for types 40 and 67, underestimates the final counts for types 19, 40, and 67, and gives a close cumulative count for the sparse type 42. This counting-process view complements the support-aware F1 analysis by showing how event time and event type predictions jointly affect accumulated fault counts.}

\subsection{\jiayurev{Multi-Day Rollout Evaluation}}
\jiayurev{The iterative generation approach allows the model to simulate system states and maintenance needs over a finite number of rollout iterations, aiding in proactive decision-making for PV plant operators. Figure \ref{supp:fig:generation} illustrates an example of rollout generation for the entire system, incorporating both event data and time series. Given the predicted next event (indicated by the red dashed line), the model generates future time series up to the expected occurrence of that event or the seven-day cap. Subsequently, the newly generated time series and event information are used to predict the next event, and this process is repeated for the specified number of iterations.}
\jiayurev{Multi-day rollout evaluation is in Section~\ref{supp:sec:rollout_evaluation}.}

\subsection{Ablation Study}
We ablate two components against the proposed multimodal model: (i)~\textbf{CLDM+MLP}, which keeps the diffusion time-series decoder but replaces SMURF-THP with an MLP event head; and (ii)~\textbf{Event-only SMURF}, which removes time-series inputs and co-attention, using event history alone.
\jiayurev{Table~\ref{tab:ablation_combined} merges overall metrics and the per-class F1 breakdowns formerly shown separately. Replacing SMURF-THP with an MLP collapses Macro/Weighted F1 from 0.3323/0.3565 to 0.0200/0.0326 and also worsens time-series MSE/MAE. Removing sensor/co-attention conditioning (Event-only) retains Macro F1 0.2627 but still trails the proposed model on aggregate metrics and on most classes with non-trivial support; Event-only remains stronger only for ultra-rare type~42 (support~5). Class-level gains are heterogeneous, so support-aware reading of the per-type rows is required.}

\begin{table}[ht]
\centering
\caption{\jiayurev{Ablation: overall metrics and per-class F1 (CLDM+MLP, Event-only, Proposed). ``--'': N/A; Supp.: test support.}}
\label{tab:ablation_combined}
\label{tab:ablation_coattention}
\jiayurev{
\setlength{\tabcolsep}{3.5pt}
\begin{tabular}{lcccc}
\toprule
\textbf{Metric / Type} & \textbf{CLDM+MLP} & \textbf{Event-only} & \textbf{Proposed} & \textbf{Supp.} \\
\midrule
\multicolumn{5}{c}{\textit{Overall}} \\
\midrule
TS MSE              & 0.4631          & --              & \textbf{0.3522} & --  \\
TS MAE              & 0.4244          & --              & \textbf{0.3418} & --  \\
Micro F1            & 0.0462          & 0.2209          & \textbf{0.3390} & 584 \\
Macro F1            & 0.0200          & 0.2627          & \textbf{0.3323} & 584 \\
Weighted F1         & 0.0326          & 0.2299          & \textbf{0.3565} & 584 \\
\midrule
\multicolumn{5}{c}{\textit{Per-class next-event F1}} \\
\midrule
4   & \textbf{0.0319} & 0.0000          & 0.0000          & 11  \\
18  & 0.0000          & 0.5000          & \textbf{0.5556} & 21  \\
19  & 0.0000          & 0.1138          & \textbf{0.2017} & 54  \\
36  & 0.0000          & 0.1250          & \textbf{0.1839} & 32  \\
40  & 0.0000          & 0.2500          & \textbf{0.5326} & 114 \\
42  & 0.0000          & \textbf{0.3636} & 0.1667          & 5   \\
53  & \textbf{0.2077} & 0.0000          & 0.1594          & 90  \\
55  & 0.0000          & 0.4662          & \textbf{0.4857} & 92  \\
67  & 0.0000          & 0.2876          & \textbf{0.4091} & 46  \\
86  & 0.0000          & 0.4762          & \textbf{0.5455} & 10  \\
95  & 0.0000          & 0.4000          & \textbf{0.4516} & 17  \\
100 & 0.0000          & 0.1695          & \textbf{0.2958} & 92  \\
\bottomrule
\end{tabular}
}
\end{table}

\noindent\textbf{Summary.} Both the SMURF-THP event decoder and multimodal co-attention are needed for the reported aggregate gains; neither ablation matches the proposed model overall.

\jiayurev{Computational efficiency analysis is in Section~\ref{supp:sec:computational_efficiency}.}
\section{Conclusions}\label{sec:conclusion}
This work presented a multi-modal generative framework for the predictive maintenance of utility-scale photovoltaic systems. By integrating an iTransformer encoder for high-dimensional sensor streams and a Transformer Hawkes Process for discrete event histories via a co-attention mechanism, our model effectively captures the complex interplay between continuous operating regimes and sporadic fault occurrences. 

A key contribution of this work is the hybrid decoder architecture: we replaced traditional deterministic forecasters with a \emph{CLDM} for sensor trajectories and a \emph{SMURF-THP} score-matching module for event timing. This design enables the generation of probabilistic, uncertainty-aware future scenarios rather than single-point estimates. Experimental evaluations on real-world data from the Red Rock Solar Site demonstrate that this approach not only achieves higher accuracy in fault prediction but also produces realistic, synchronized rollouts of sensor data and maintenance events.

Future work includes generalization, deployment, and RL-based maintenance policies, with the model as a digital-twin sandbox for offline RL and off-policy evaluation \cite{wang2022digitaltwinconveyor,xie2025windturbinerl}.

\ifblind
\else
\section*{Acknowledgment}
The research reported in this paper is supported by funds from the U.S. Department of Energy's Office of Energy Efficiency and Renewable Energy (EERE) under the Solar Energy Technology Office (SETO) Award Number DE-EE0009354 (Program Manager: Dr. Tassos Golnas). The support is gratefully acknowledged.
\fi

\bibliographystyle{IEEEtran}
\bibliography{ref}

\clearpage
\setcounter{section}{0}
\setcounter{figure}{0}
\setcounter{table}{0}
\setcounter{equation}{0}
\renewcommand{\thesection}{S\arabic{section}}
\renewcommand{\thesubsection}{S\arabic{section}.\arabic{subsection}}
\renewcommand{\thefigure}{S\arabic{figure}}
\renewcommand{\thetable}{S\arabic{table}}
\renewcommand{\theequation}{S\arabic{equation}}

\section*{Supplementary Material}
\addcontentsline{toc}{section}{Supplementary Material}

\jiayurev{This supplementary material provides details of the rollout procedure and evaluation, additional visualizations, experimental settings, and supporting methodological and data descriptions.}

\section{Finite-Step Rollout Generation}
\label{supp:sec:rollout_generation}
\begin{figure}[htbp]
  \centering
  \includegraphics[width=0.8\columnwidth]{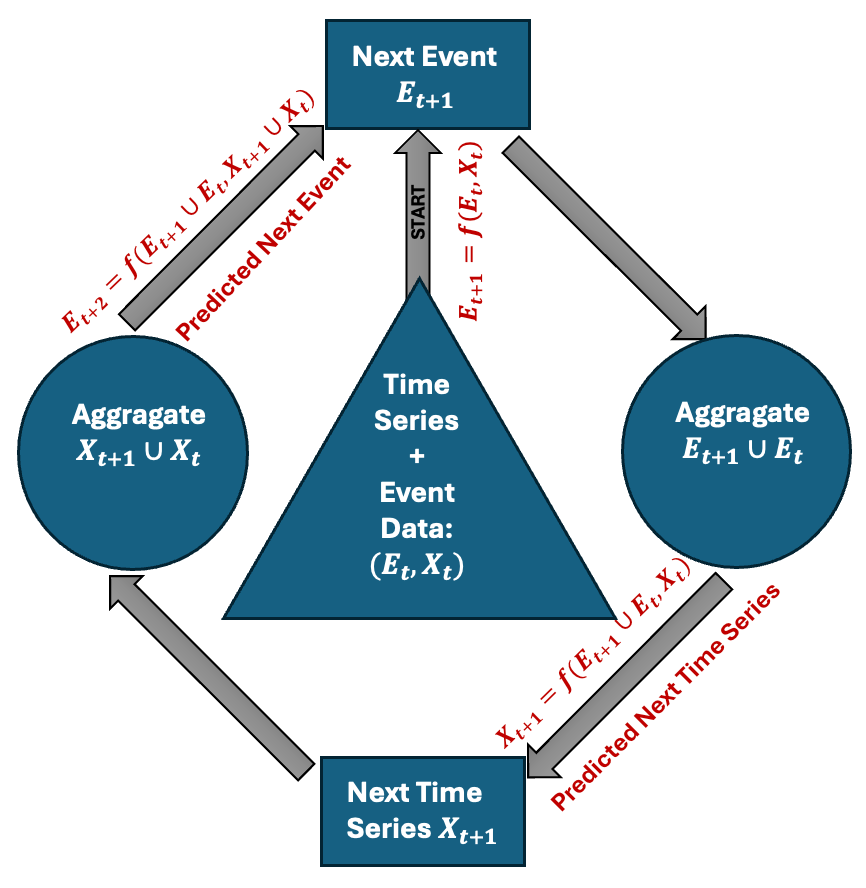}
  \caption{\jiayurev{Finite-step rollout structure. At each iteration, only the available history (context window of time series and last $K$ events up to current time $t$) is used to predict the next event and to generate the time series segment up to the predicted event time, capped by the seven-day prediction window; then the window is rolled forward. The number of generation iterations $N_{\mathrm{iter}}$ is specified, while the calendar duration is induced by generated inter-event intervals and the per-iteration seven-day cap.}}
  \label{supp:fig:generation_method}
\end{figure}

\begin{figure}[htbp]
  \centering
\includegraphics[width=1\linewidth]{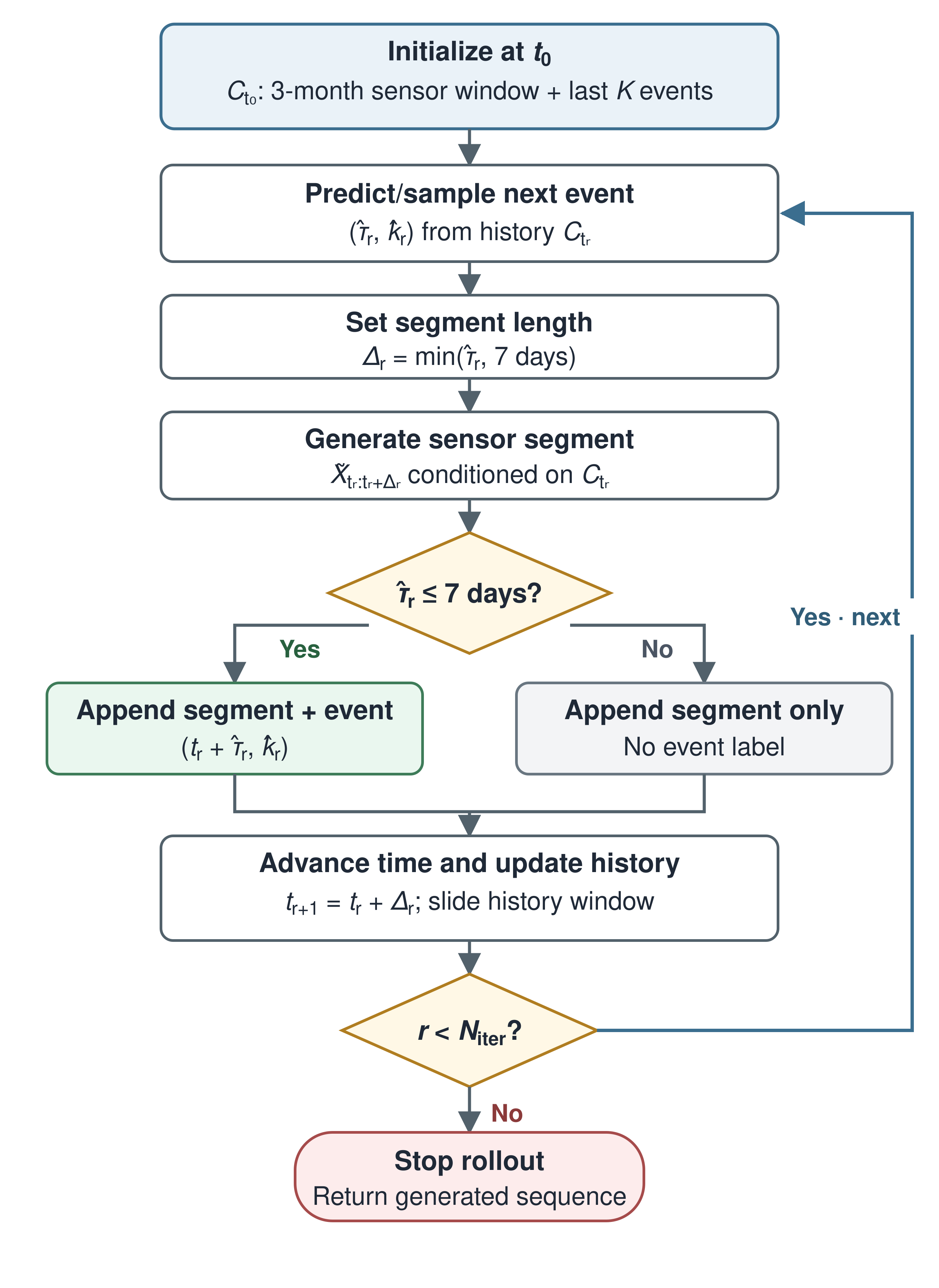}
\caption{\jiayurev{Supplementary rollout flowchart. The rollout initializes from the available history, predicts or samples the next event, generates the sensor segment capped by the seven-day prediction window, updates the history, and repeats until the specified number of generation iterations $N_{\mathrm{iter}}$ is reached.}}
  \label{supp:fig:generation_flowchart}
\end{figure}

\jiayurev{The finite-step rollout procedure is specified through the available history and the chronological update steps below, with its structure and workflow illustrated in Figures~\ref{supp:fig:generation_method} and~\ref{supp:fig:generation_flowchart}, respectively.}

\paragraph{Available history at time $t$.}
At any time $t$, the model's inputs are limited to a fixed \emph{context window} of past data: (i) a segment of observed (or previously generated) time series over a lookback horizon (e.g., the last few months up to $t$), and (ii) the last $K$ events with occurrence times $\le t$. No information after $t$ is used.

\paragraph{\jiayurev{Chronological rollout iterations}}
\begin{itemize}
    \item \textbf{Initialize:} Set the current time to the end of the observed training/validation period. The available history is the observed context window at that time (e.g., last 3 months of time series and last 10 events up to that time). All subsequent steps use only this history and whatever has been generated so far.
    
    \item \textbf{Predict next event:} Given the current available history (time series and events up to the current time), the model predicts the next event---its time $\hat{t}_{\text{next}}$ and type. This prediction is based solely on the encoder's representation of the past; no future data are used.
    
    \item \textbf{Generate time series until $\hat{t}_{\text{next}}$ or the prediction-window limit:} Conditioning on the same history (and optionally the predicted event time/type), the diffusion decoder generates a trajectory of multivariate sensor data from the current time up to $\hat{t}_{\text{next}}$, capped by the seven-day prediction window. The length of this segment is determined by the predicted inter-event interval and the prediction-window cap, not by any future ground truth.
    
    \item \textbf{Roll the window forward:} Append the generated time-series segment to the running history. If $\hat{t}_{\text{next}}$ falls within the seven-day prediction window, also append the predicted event $(\hat{t}_{\text{next}}, \hat{k})$; otherwise, no event label is recorded for that rollout iteration. Advance the current time by the generated segment length and truncate or slide the context window so that it again contains only the last lookback period of time series and the last $K$ available events. Repeat from ``Predict next event'' for the next iteration until $N_{\mathrm{iter}}$ iterations are completed.
\end{itemize}
\jiayurev{This update rule is forward-in-time and history-conditioned: each prediction and each generated segment depend only on data up to the current time, avoiding any leakage from future observations.}

\section{Multi-Day Rollout Evaluation}
\label{supp:sec:rollout_evaluation}
\jiayurev{The rollout procedure is applied to the proposed model and is stochastic in both modalities. At each rollout iteration, the sensor segment is sampled from the CLDM reverse process, the next event time is sampled through the SMURF-THP score-based procedure, and the next event type is sampled from the learned categorical distribution. Deterministic summaries, such as the sample mean trajectory or the maximum-probability event type, are used only for point-metric reporting.}

\begin{figure}[ht]
  \centering
  \includegraphics[width=\columnwidth]{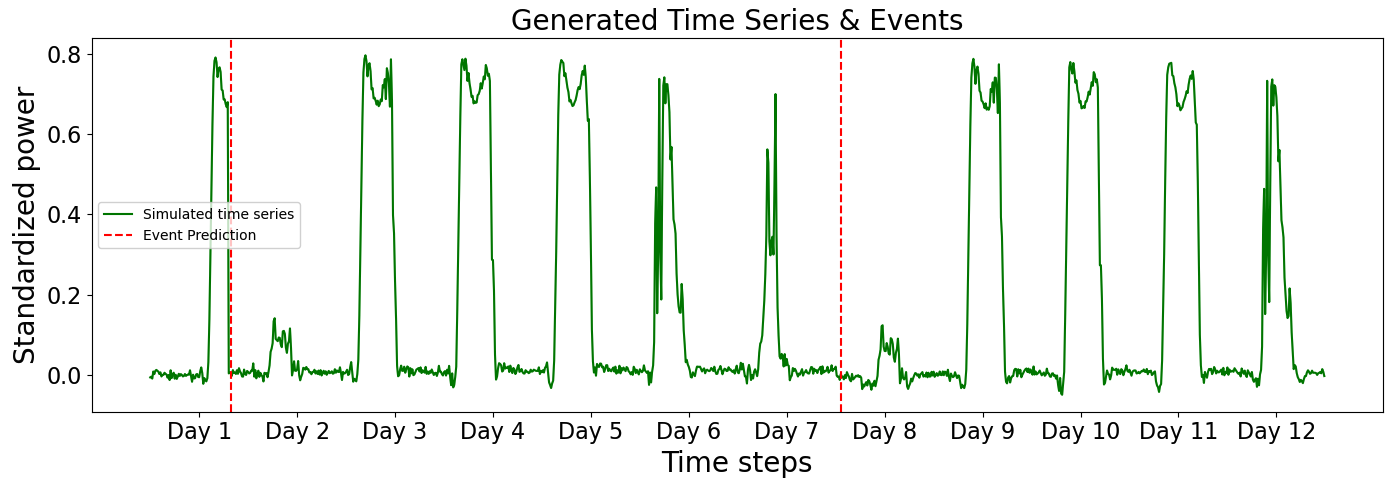}
  \caption{\jiayurev{Multi-day rollout visualization showing iterative event prediction and time series generation.}}
  \label{supp:fig:generation}
\end{figure}

\jiayurev{In this case study, the visualization uses a specified number of rollout iterations $N_{\mathrm{iter}}=8$. Because each iteration is capped by the seven-day prediction window, the realized calendar span is determined by the generated inter-event intervals and the per-iteration cap. As $N_{\mathrm{iter}}$ increases, stochastic event timing and sensor evolution can make individual paths diverge from the realized trajectory, so the rollout is intended for risk-sensitive scenario analysis. We therefore summarize longer rollouts through distributional quantities such as mean trajectories, prediction intervals, event-count envelopes, and cumulative counting-process errors, rather than requiring exact pathwise agreement for every sampled scenario.}

\begin{figure}[htbp]
  \centering
  \includegraphics[width=\columnwidth]{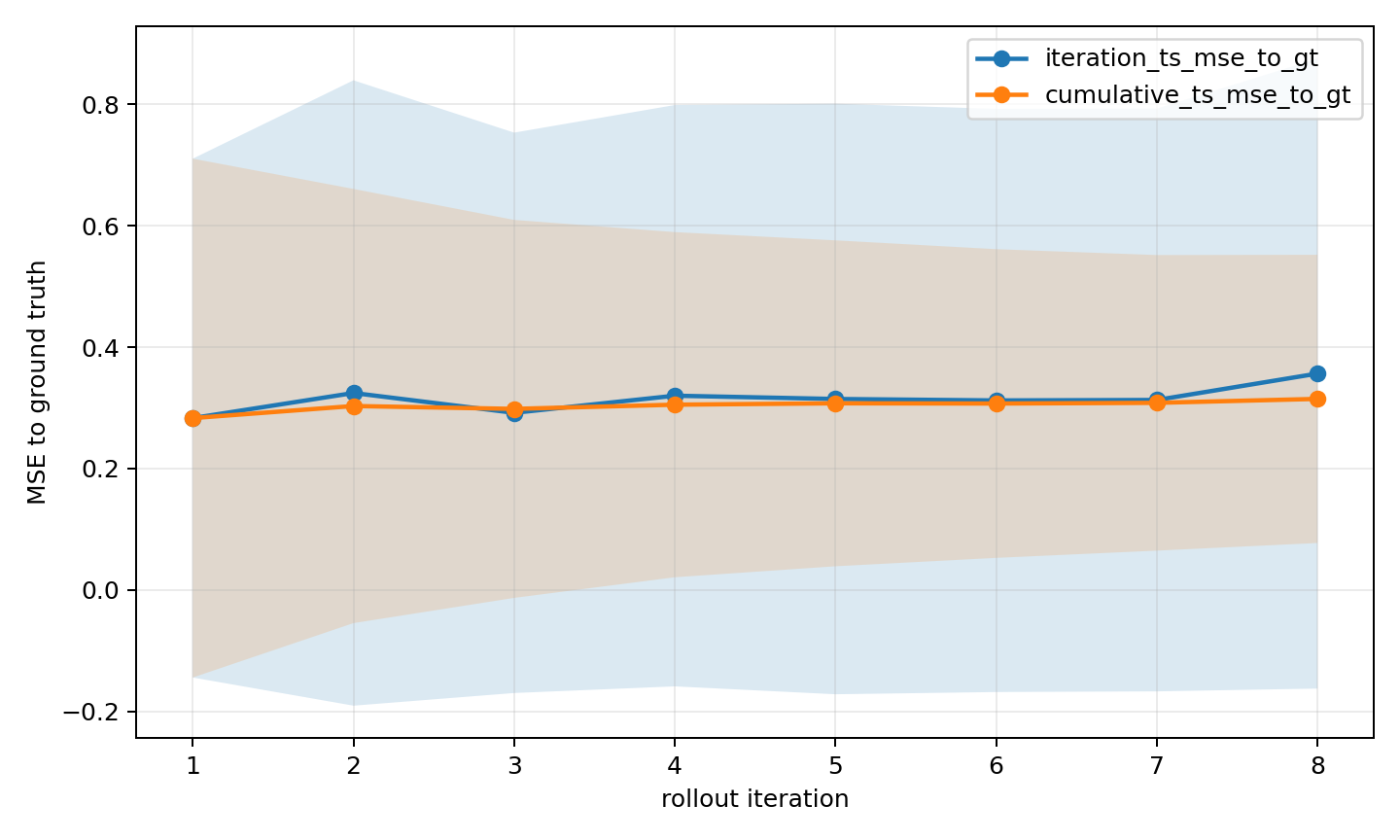}
  \caption{\jiayurev{Multi-step rollout error accumulation over $N_{\mathrm{iter}}=8$ iterations. The blue curve shows per-iteration MSE and the orange curve shows cumulative MSE relative to ground truth; shaded regions denote the standard deviation across samples. Cumulative MSE remains stable near $0.30$ without exponential growth, indicating controlled trajectory drift under event-driven alignment.}}
  \label{supp:fig:error_accumulation}
\end{figure}
\jiayurev{To quantitatively evaluate stability and error accumulation during multi-step generation, Figure~\ref{supp:fig:error_accumulation} reports per-iteration and cumulative MSE over eight rollout iterations. The cumulative MSE remains highly stable at approximately $0.30$, while the per-iteration MSE shows only bounded fluctuations (about $0.28$--$0.36$) without exponential growth. This supports the claim that the event-driven, history-conditioned rollout mitigates the compounding drift typical of purely autoregressive long-horizon forecasting.}

\section{Additional Power-Signal Forecast Visualizations}
\label{supp:app:more_power_vis}
\begin{figure*}[h]
  \centering
  \includegraphics[width=\textwidth]{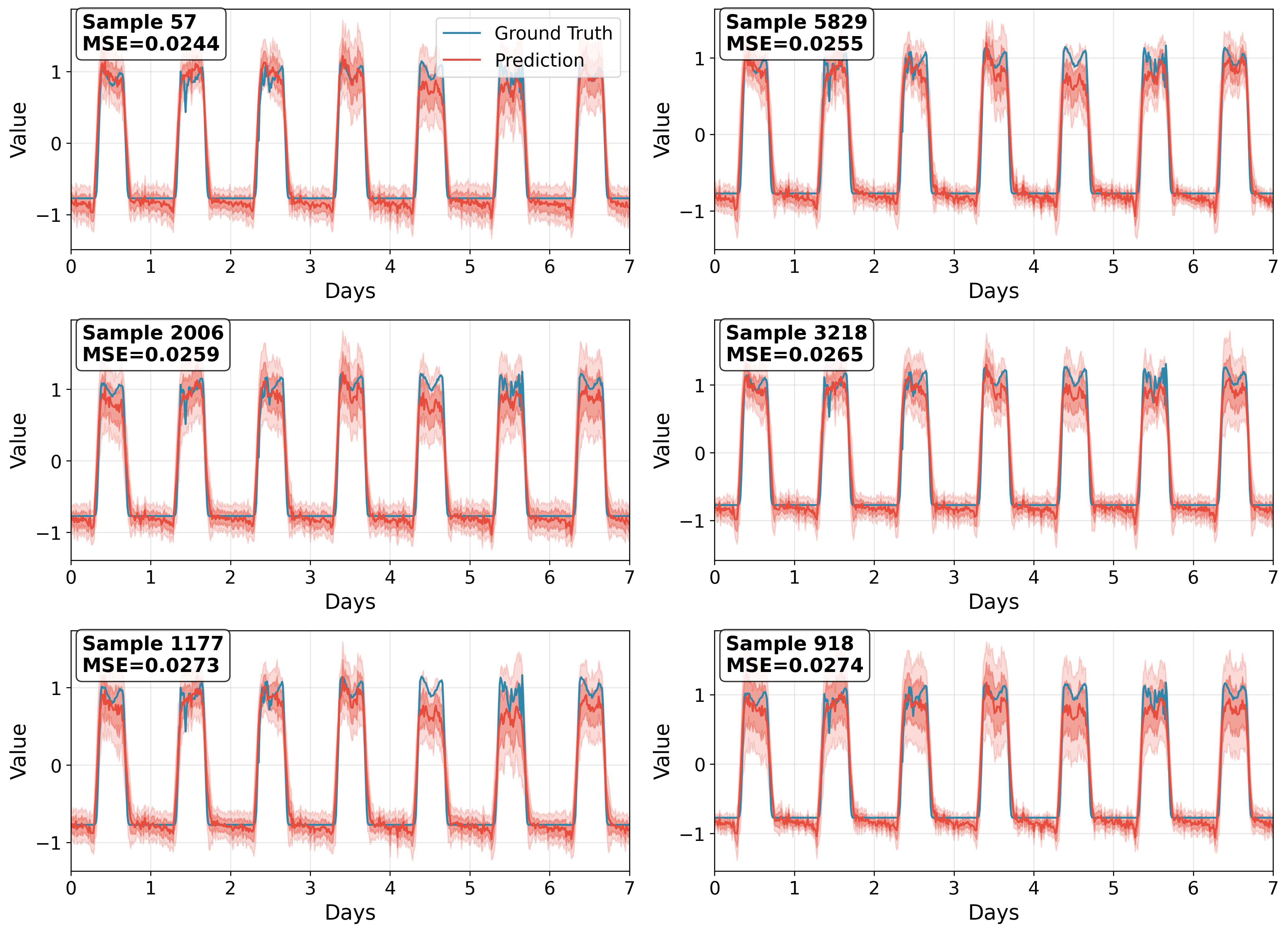}
  \caption{\jiayurev{Additional seven-day CLDM power-signal generation examples on six randomly selected test samples. Blue: ground truth; red: predicted mean; shaded red: sample-based uncertainty bands. Each panel reports the sample index and MSE.}}
  \label{supp:fig:more_power_visualization_mse}
\end{figure*}

\jiayurev{Figure~\ref{supp:fig:more_power_visualization_mse} complements the representative seven-day power forecast in Figure~\ref{fig:ts_and_next_event} and the finite-step rollout discussion by showing six randomly selected test samples. In each panel, the blue curve is the ground-truth power signal, the red curve is the CLDM mean prediction, and the shaded red bands are sample-based predictive uncertainty intervals from repeated diffusion draws. The panels show that generated trajectories preserve the diurnal on/off PV pattern, track daytime fluctuations, and keep night-time levels in the stable low-output regime, while the annotated per-sample MSE values remain low (approximately $0.024$--$0.027$). These examples support interpreting the diffusion decoder as providing high-fidelity short-horizon reconstructions with usable uncertainty bands for maintenance-oriented scenario analysis.}

\begin{figure}[htbp]
  \centering
  \includegraphics[width=0.92\columnwidth]{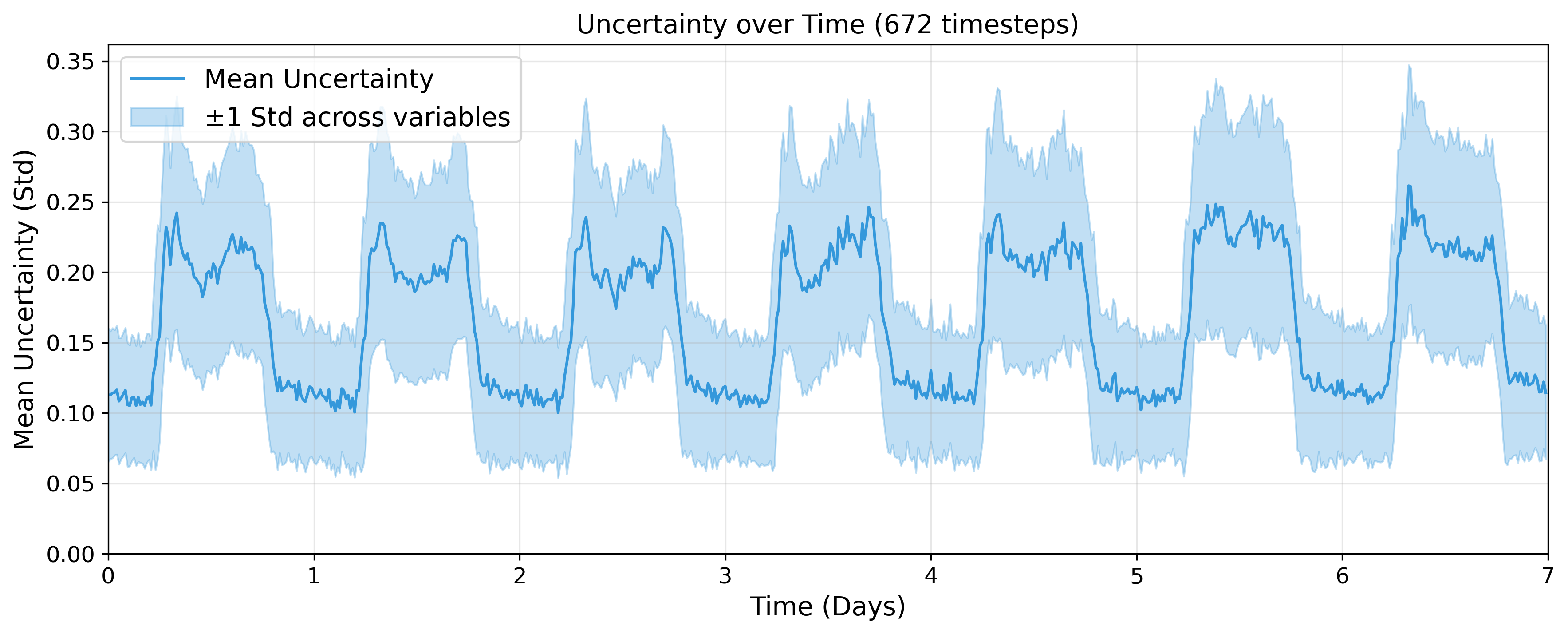}
  \caption{Mean predictive uncertainty (averaged across variables) over a 7-day horizon on the test set.}
  \label{supp:fig:uncertainty_dynamics}
\end{figure}
\jiayurev{Figure~\ref{supp:fig:uncertainty_dynamics} aggregates the sample standard deviation of CLDM trajectories across the test set. Uncertainty is heteroscedastic and diurnal: larger during morning ramp-up and smaller during evening decline and nighttime.}

\section{Tuning Parameters}
Random search is used for hyperparameter tuning, with the search ranges shown below:

\begin{table}[h]
\centering
\caption{Common hyperparameter settings for all models.}
\label{supp:tab:common_hyperparameters}
\footnotesize
\setlength{\tabcolsep}{4pt}
\renewcommand{\arraystretch}{1.05}
\begin{tabular}{p{0.58\columnwidth}p{0.30\columnwidth}}
\hline
\textbf{Parameter} & \textbf{Range} \\
\hline
Learning rate
& $[5\times10^{-5},\,3\times10^{-4}]$ \\

Number of event types
& 14 \\

Training sampler
& Event-balanced \\

Batch size
& 128 \\

Maximum epochs
& $\{80,100,120\}$ \\

Early-stopping patience
& $\{15,25\}$ \\
\hline
\end{tabular}
\end{table}

\begin{table}[h]
\centering
\caption{Hyperparameter search space for the proposed model.}
\label{supp:tab:proposed_hyperparameters}
\begin{tabular}{ll}
\hline
Hyperparameter & Candidate values \\
\hline
Time-series loss weight, $w_{\mathrm{ts}}$
    & $\{0.5, 0.7, 1.0\}$ \\
Projection loss weight, $w_{\mathrm{proj}}$
    & $\{1.0, 1.5, 2.0\}$ \\
Number of latent diffusion steps
    & $\{100, 200\}$ \\
Event classification loss weight, $w_{\mathrm{event}}$
    & $\{0.5, 1.0, 2.0\}$ \\
Next-event loss weight, $w_{\mathrm{next}}$
    & $\{0.05, 0.1, 0.2\}$ \\
Focal-loss parameter, $\gamma$
    & $\{0.0, 1.0, 2.0\}$ \\
Event-class weighting
    & inverse square root, inverse \\
Sample-weight exponent, $\alpha$
    & $\{0.5, 0.75, 1.0\}$ \\
Maximum sample weight
    & $\{5.0, 10.0\}$ \\
Event encoder hidden dimension
    & $\{128, 256\}$ \\
\hline
\end{tabular}
\end{table}

\begin{table}[h]
\centering
\caption{Hyperparameter search space for the THP baseline.}
\label{supp:tab:thp_hyperparameters}
\footnotesize
\setlength{\tabcolsep}{4pt}
\renewcommand{\arraystretch}{1.05}
\begin{tabular}{p{0.58\columnwidth}p{0.30\columnwidth}}
\hline
\textbf{Parameter} & \textbf{Range} \\
\hline
Learning rate
& $[5\times10^{-5},\,3\times10^{-4}]$ \\

Event loss weight, $w_{\mathrm{event}}$
& $\{0.5,1.0,2.0\}$ \\

Next-event loss weight, $w_{\mathrm{next}}$
& $\{0.05,0.1,0.2\}$ \\

Focal-loss parameter, $\gamma$
& $\{0,1,2\}$ \\

Event-class weighting
& \{Inv. sqrt., Inv.\} \\

Sample-weight exponent, $\alpha$
& $\{0.5,0.75,1.0\}$ \\

Maximum sample weight
& $\{5,10\}$ \\

Hidden dimension
& $\{128,256\}$ \\

Transformer layers
& $\{2,3\}$ \\

Attention heads
& $\{4,8\}$ \\
\hline
\end{tabular}
\end{table}

\begin{table}[h]
\centering
\caption{Hyperparameter search space for the CLDM+MLP baseline.}
\label{supp:tab:cldm_mlp_hyperparameters}
\footnotesize
\setlength{\tabcolsep}{4pt}
\renewcommand{\arraystretch}{1.05}
\begin{tabular}{p{0.58\columnwidth}p{0.30\columnwidth}}
\hline
\textbf{Parameter} & \textbf{Range} \\
\hline
Time-series loss weight, $w_{\mathrm{ts}}$
& $\{0.5,0.7,1.0\}$ \\

Projection loss weight, $w_{\mathrm{proj}}$
& $\{1.0,1.5,2.0\}$ \\

Latent diffusion steps
& $\{100,200\}$ \\

Event loss weight, $w_{\mathrm{event}}$
& $\{0.5,1.0,2.0\}$ \\

Next-event loss weight, $w_{\mathrm{next}}$
& $\{0.05,0.1,0.2\}$ \\

Focal-loss parameter, $\gamma$
& $\{0,1,2\}$ \\

Event-class weighting
& \{Inv. sqrt., Inv.\} \\

Sample-weight exponent, $\alpha$
& $\{0.5,0.75,1.0\}$ \\

Maximum sample weight
& $\{5,10\}$ \\
\hline
\end{tabular}
\end{table}

\begin{table}[h]
\centering
\caption{Hyperparameter search space for the event-only model.}
\label{supp:tab:event_only_hyperparameters}
\footnotesize
\setlength{\tabcolsep}{4pt}
\renewcommand{\arraystretch}{1.05}
\begin{tabular}{p{0.58\columnwidth}p{0.30\columnwidth}}
\hline
\textbf{Parameter} & \textbf{Range} \\
\hline
Time-series loss weight, $w_{\mathrm{ts}}$
& 0 (fixed) \\

Event loss weight, $w_{\mathrm{event}}$
& $\{0.5,1.0,2.0\}$ \\

Next-event loss weight, $w_{\mathrm{next}}$
& $\{0.05,0.1,0.2\}$ \\

Focal-loss parameter, $\gamma$
& $\{0,1,2\}$ \\

Event-class weighting
& \{Inv. sqrt., Inv.\} \\

Sample-weight exponent, $\alpha$
& $\{0.5,0.75,1.0\}$ \\

Maximum sample weight
& $\{5,10\}$ \\

Event encoder hidden dimension
& $\{128,256\}$ \\
\hline
\end{tabular}
\end{table}

\begin{table}[h]
\centering
\caption{Hyperparameter search space for the RNNPP baseline.}
\label{supp:tab:rnnpp_hyperparameters}
\footnotesize
\setlength{\tabcolsep}{4pt}
\renewcommand{\arraystretch}{1.05}
\begin{tabular}{p{0.58\columnwidth}p{0.30\columnwidth}}
\hline
\textbf{Parameter} & \textbf{Range} \\
\hline
Time-series loss weight, $w_{\mathrm{ts}}$
& 0 (fixed) \\

Event loss weight, $w_{\mathrm{event}}$
& $\{0.5,1.0,2.0\}$ \\

Next-event loss weight, $w_{\mathrm{next}}$
& $\{0.05,0.1,0.2\}$ \\

Focal-loss parameter, $\gamma$
& $\{0,1,2\}$ \\

Event-class weighting
& \{Inv. sqrt., Inv.\} \\

Sample-weight exponent, $\alpha$
& $\{0.5,0.75,1.0\}$ \\

Maximum sample weight
& $\{5,10\}$ \\

Event encoder hidden dimension
& $\{128,256\}$ \\

Time-series encoder hidden dimension
& $\{64,128\}$ \\

Fusion dimension
& $\{128,256\}$ \\

Event context length
& 96 (fixed) \\
\hline
\end{tabular}
\end{table}

\section{Comparison of Representative Methods}
\label{supp:sec:method_comparison}
\begin{table*}[t]
\centering
\caption{Representative methods for event diffusion/flow, joint event--time-series co-prediction, and time-series forecasting (Transformers and diffusion/score-based).}
\label{supp:tab:temporal-gen-comparison}
\footnotesize
\setlength{\tabcolsep}{3pt}
\renewcommand{\arraystretch}{1.08}
\begin{tabular}{@{}p{0.34\linewidth}p{0.22\linewidth}p{0.40\linewidth}@{}}
\toprule
Method & Conditioning & Scope / design \\
\midrule
\multicolumn{3}{@{}l}{\textbf{Event sequences (TPP): diffusion / flow}}\\
ADD-THIN \cite{ludke2023addthin} &
Event history &
Full-sequence denoising via add-and-thin; handles discrete+continuous components \\
DLTPP \cite{zhou2023dltpp} &
History (mapped) &
Non-autoregressive long-horizon; bidirectional map between sequences and Euclidean space \\
Zeng et al. (CDiff) \cite{zeng2024interacting} &
Event history &
Interacting diffusion processes for inter-arrival times and types (sequence-to-sequence) \\
ADiff4TPP \cite{mukherjee2025adiff4tpp} &
Event history &
Asynchronous noise schedule to strengthen long-horizon conditioning \\
EventFlow \cite{kerrigan2026eventflow} &
Event history &
Flow-matching-based forecasting for continuous-time event sequences \\
SMURF-THP \cite{li2023smurf} &
THP context &
Score-matching UQ; fixed-noise-level denoising for next-event timing \\
\textbf{Proposed} &
Cross-modal &
Next-event score matching on $\log\tau$; type head; jointly rolled out with TS diffusion decoder \\
\midrule
\multicolumn{3}{@{}l}{\textbf{Joint event + time series (co-prediction)}}\\
RPPN (AAAI) \cite{xiao2017modeling} &
Events + time series &
Point-process intensity with dual RNNs (time-series background + event history) \\
Attentional Twin-RNN \cite{xiao2017joint} &
Events + time series &
Co-evolving encoders + attention; collaborative modeling of events and series \\
RPPN (TNNLS) \cite{xiao2019learning} &
Time series $\leftrightarrow$ events &
Collaborative learning of time-series-associated event sequences \\
Generative RNN-ODE \cite{li2022learning} &
Events + time series &
Continuous-time latent dynamics for collaborative forecasting/generation \\
\midrule
\multicolumn{3}{@{}l}{\textbf{Time series forecasting: Transformers}}\\
Informer \cite{zhou2021informer} &
Past window (+ covariates) &
Efficient long-horizon Transformer via ProbSparse attention \\
Autoformer \cite{wu2021autoformer} &
Past window &
Decomposition + auto-correlation mechanism for long-term forecasting \\
PatchTST \cite{nie2023patchtst} &
Past window &
Patch-based / channel-independent Transformer backbone \\
TimesNet \cite{wu2023timesnet} &
Past window &
Multi-periodicity modeling (temporal 2D-variation) for general TS tasks \\
iTransformer \cite{liu2023itransformer} &
Past window &
Inverted tokenization (variate tokens); strong long-lookback multivariate TS \\
\midrule
\multicolumn{3}{@{}l}{\textbf{Time series forecasting/generation: diffusion / score-based}}\\
TimeGrad \cite{rasul2021autoregressive} &
Past window &
Autoregressive diffusion/score sampling for probabilistic forecasts \\
CSDI \cite{tashiro2021csdi} &
Observations / masks &
Conditional score diffusion (imputation; can be used for forecasting) \\
TimeDiff \cite{shen2023timediff} &
Past window &
Conditional non-autoregressive diffusion for long-horizon prediction \\
TSDiff \cite{kollovieh2023tsdiff} &
Unconditional + guidance &
Unconditional diffusion + self-guidance for forecasting/refinement/synthesis \\
\bottomrule
\end{tabular}
\end{table*}

\section{iTransformer Schematic}
\label{supp:sec:itransformer_schematic}
\begin{figure}[htbp]
  \centering
\includegraphics[width=0.9\columnwidth]{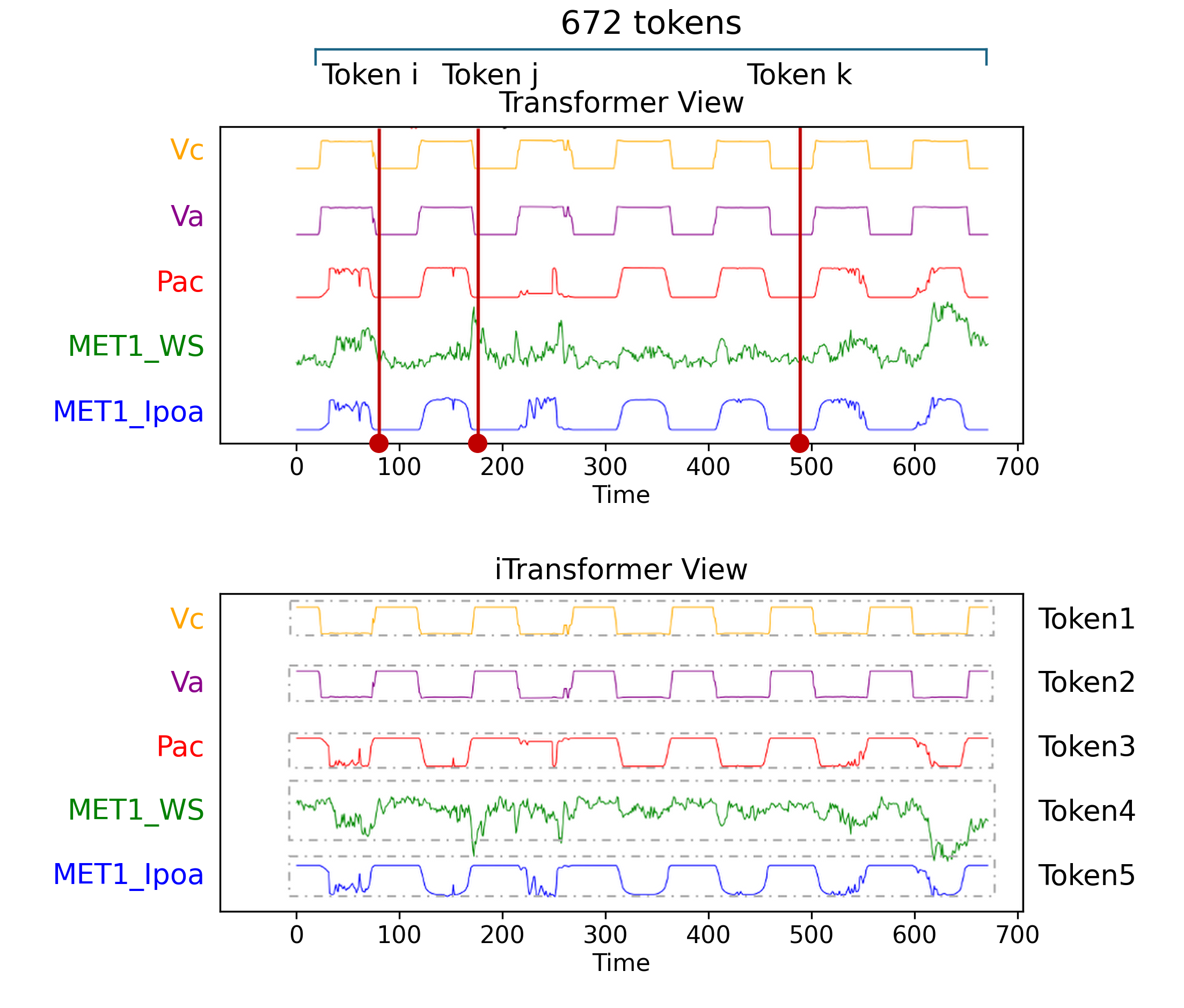}
  \caption{Comparison between vanilla Transformer (top) and the iTransformer (bottom): normalized time series from five randomly chosen sensors over 7 days, segmented into 96 time intervals per day. The iTransformer treats each variate as a token and applies attention across sensors.}
  \label{supp:fig:itrans}
\end{figure}

\section{Co-Attention Layer Architecture}
\label{supp:sec:coattention_architecture}
\begin{figure}[htbp]
  \centering
\includegraphics[width=0.82\columnwidth]{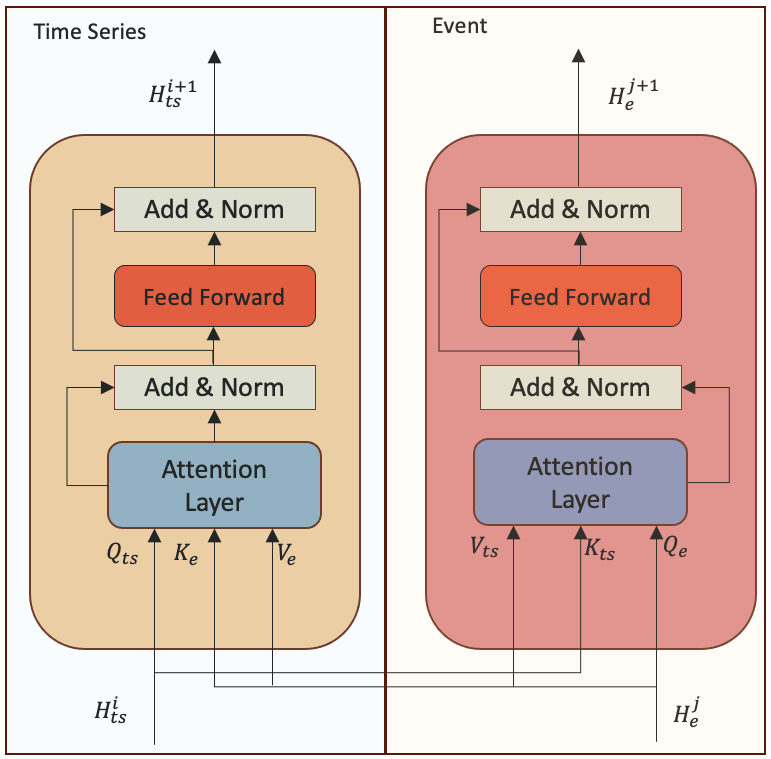}
  \caption{Co-attention Layer Architecture \cite{lu2019vilbert}. The bidirectional cross-attention mechanism enables each modality to attend to the other, capturing cross-modal interdependencies.}
  \label{supp:fig:coattention}
\end{figure}

\section{Fault Code Descriptions}
\label{supp:sec:fault_code_descriptions}
\begin{table}[htbp]
\centering
\caption{Fault Codes' Descriptions}
\label{supp:t:fault_descriptions}
\begin{tabular}{ll}
\toprule
\textbf{Fault type} & \textbf{Description} \\
\midrule
4 & Over current S/W \\
12 & Multiple issues \\
18 & Multiple issues \\
19 & AC Overcurrent issue \\
36 & High grid frequency \\
40 & Homopolar current Fault \\
42 & Multiple issue \\
53 & Leakage current anomaly \\
55 & Switch doesn't work in AC main contactor \\
67 & Differential voltage anomaly \\
86 & IGBT temperature exceeded threshold \\
95 & Photovoltaic generation anomaly \\
100 & Other infrequent faults \\
\bottomrule
\end{tabular}
\end{table}

\section{Processed Fault Data Example}
\begin{table}[ht]
\centering
\caption{Processed Real-Time Fault Data Sample}
\setlength{\tabcolsep}{4pt}
\label{supp:t:real_time_fault_data}
\begin{tabular}{lccc}
\hline
\textbf{Event Start Time} & \textbf{Gap Time (mins)} & \textbf{Event Type} & \textbf{Duration (mins)} \\
\hline
2019-06-24 11:53:10 & 0 & 55 & 17.06 \\
2019-06-24 14:01:18 & 128.13 & 18 & 401.73 \\
2019-06-30 18:18:14 & 8896.93 & 55 & 1736.66 \\
2019-07-02 05:29:34 & 2111.33 & 55 & 3979.33 \\
2019-07-05 05:29:26 & 4319.86 & 55 & 1065.73 \\
2019-07-06 05:41:50 & 1452.39 & 55 & 2406.93 \\
\bottomrule
\end{tabular}
\end{table}

\section{THP Encoder Formulation}
\label{supp:sec:thp_encoder_formulation}
For different event types, e.g., $k \in [1,2,\ldots,K]$, this method defines a continuous conditional intensity function for event type $k$ as \cite{zuo2020transformer}:
\begin{equation}
    \lambda_k(t \mid \mathcal{H}_{t_j})=f_k\left(\alpha_k(t-t_j)+\mathbf{w}_k^\top \mathbf{h}(t_j)+b_k\right), \quad t>t_j,
\end{equation}
where $t_j$ is the time of the most recent event before $t$, and the history $\mathcal{H}_{t_j}=\{(t_i,k_i):t_i \le t_j\}$ is the set of events up to and including $t_j$. The link function $f_k(\cdot)$ is the softplus: $f_k(\cdot)=\mathrm{softplus}(\cdot)$ (in \cite{zuo2020transformer}, the parameterized form $\beta_k\log(1+\exp(\cdot/\beta_k))$ is used for numerical stability). The terms are:
\begin{itemize}
    \item $\alpha_k(t-t_j)$ is the current-time term (zero at $t=t_j$, increasing with elapsed time since $t_j$).
    \item $\mathbf{w}_k^\top \mathbf{h}(t_j)$ is the history term from the hidden state at $t_j$.
    \item $b_k$ is the base intensity for type $k$.
\end{itemize}
Together, these components allow the model to capture complex event dynamics by combining current temporal influence, historical dependencies, and base event likelihood.
\jiayurev{The modeling assumptions are as follows. The event log is treated as a marked temporal point process observed through a finite history window, and the next-event distribution is assumed to be conditionally determined by the event history representation and the synchronized sensor context. The event mark is the fault type only; duration is recorded in the raw log but is not used as a mark in the event decoder. We do not assume a globally stationary fault process. Instead, changes in operating regime are represented through the time-series encoder and the fused context $\mathbf{h}$, so the conditional distribution of the next event can vary with recent sensor behavior.}

For predicting the next event time stamp $\hat{t}_{j+1}$ and next event type $\hat{k}_{j+1}$, the following equations are used:
\begin{align}
    & p(t|\mathcal{H}_t) = \lambda(t|\mathcal{H}_t)\exp\left(-\int_{t_j}^{t} \lambda(\tau|\mathcal{H}_{\tau}) \, d\tau\right) \\
    & \hat{t}_{j+1} = \int_{t_j}^{\infty} t \cdot p(t|\mathcal{H}_t) \, dt \\
    & \hat{k}_{j+1} = \argmax_{k} \frac{\lambda_k(t_{j+1}|\mathcal{H}_{j+1})}{\lambda(t_{j+1}|\mathcal{H}_{j+1})}
\end{align}

In these equations, $p(t|\mathcal{H}_t)$ represents the probability density function of an event that occurs exactly at time $t$ given the history $\mathcal{H}_t$. The next event time $\hat{t}_{j+1}$ is calculated as the expected value of $t$ in the distribution $p(t|\mathcal{H}_t)$, while the next type of event $\hat{k}_{j+1}$ is determined by maximizing the ratio of the type-specific intensity $\lambda_k(t_{j+1}|\mathcal{H}_{j+1})$ to the overall intensity $\lambda(t_{j+1}|\mathcal{H}_{j+1})$. \jiayurev{These THP equations describe the intensity-based encoder and event-only baseline. In the proposed rollout, event timing is generated by the SMURF-THP score model in log-inter-event-time space, and event type is sampled from the learned categorical distribution for stochastic scenarios; the argmax type is used only when a single deterministic label is needed for F1-score reporting.}

\section{Computational Efficiency}
\label{supp:sec:computational_efficiency}
\jiayurev{Let $C$ be the number of variables, $L$ the time-series lookback length, $M$ the event-context length, $d$ the hidden dimension, and $S$ the number of diffusion denoising steps. After the lookback embedding, the iTransformer encoder applies attention over $C$ variate tokens, giving an attention cost of approximately $O(C^2d)$; the THP event encoder costs $O(M^2d)$ for masked event self-attention. The CLDM decoder cost scales linearly with $S$ and with the number of sampled trajectories used for uncertainty estimation. In this case study, $C=61$ and $M=10$, so the dominant cost is the conditional time-series diffusion decoder rather than the event encoder.}

\jiayurev{This complexity profile also explains the practical design trade-off in the hybrid decoder. The event module is kept as a lightweight score-matching next-event decoder, while the higher-capacity iterative diffusion process is reserved for high-dimensional sensor trajectories where distributional uncertainty is most expensive to represent. In applications that require faster online inference, the number of diffusion steps or sampled trajectories can be reduced, whereas offline maintenance planning can use more samples to obtain tighter uncertainty estimates.}

\jiayurev{To assess practical deployability, we evaluated the framework's computational footprint on the case study dataset. As summarized in Table \ref{supp:tab:comp_efficiency}, the model consists of approximately 171.1 million trainable parameters. It exhibits an efficient memory footprint, utilizing a peak GPU memory of 8,241 MiB, which allows for deployment on standard hardware without multi-GPU clusters. The temporal cost is also scalable, averaging roughly 72 seconds per epoch, enabling the full 120-epoch training cycle to complete in under 2.5 hours.

\begin{table}[htbp]
\centering
\caption{Computational Efficiency and Model Footprint}
\label{supp:tab:comp_efficiency}
\jiayurev{\begin{tabular}{@{}ll@{}}
\toprule
\textbf{Metric} & \textbf{Value} \\ 
\midrule
Total / Trainable Parameters & 171,086,433 ($\sim$171.1 M) \\
Peak GPU Memory Footprint & 8,241 MiB ($\sim$8.2 GB) \\
Training Time per Epoch & 1m 11.97s \\
Total Epochs & 120 \\
Estimated Total Training Time & $\sim$2 hours 24 minutes \\
\bottomrule
\end{tabular}}
\end{table}
}

\section{Reproducibility Settings}
\begin{table}[ht]
\centering
\caption{\jiayurev{Reproducibility settings fixed in the case study}}
\label{supp:tab:reproducibility_settings}
\setlength{\tabcolsep}{4pt}
\begin{tabular}{ll}
\toprule
\textbf{Item} & \textbf{Setting} \\
\midrule
Event context length & 10 previous events \\
Time-series lookback & 3 months \\
\jiayurev{Prediction window} & \jiayurev{7 days after forecast origin} \\
Event-time variable & $x=\log \tau$ \\
\jiayurev{Loss weights} & \jiayurev{$0.7(\mathcal{L}_{\mathrm{diff}}+1.5\mathcal{L}_{\mathrm{proj}})+2.0\mathcal{L}_{\mathrm{event}}+0.1\mathcal{L}_{\mathrm{time}}$} \\
\jiayurev{CLDM samples / steps} & \jiayurev{20 samples, 50 diffusion steps (uncertainty eval.)} \\
\bottomrule
\end{tabular}
\end{table}

\end{document}